\documentclass[10pt,journal,compsoc]{IEEEtran}

\usepackage[T1]{fontenc}
\usepackage[utf8]{inputenc}
\usepackage{microtype}        

\usepackage{amsmath}
\usepackage{amssymb}
\usepackage{bm}               

\usepackage{booktabs}         
\usepackage{multirow}
\usepackage{makecell}         
\usepackage{array}
\usepackage{tabularx}
\usepackage{rotating}         

\usepackage{graphicx}

\usepackage[table]{xcolor}
\definecolor{lightyellow}{RGB}{255,255,204}
\definecolor{lightgreen}{RGB}{204,255,204}
\definecolor{lightred}{RGB}{255,204,204}
\definecolor{lightblue}{RGB}{204,229,255}
\definecolor{IEEEblue}{RGB}{0,82,155}

\usepackage[
  colorlinks=true,
  linkcolor=IEEEblue,
  citecolor=IEEEblue,
  urlcolor=IEEEblue,
  bookmarksopen=true
]{hyperref}
\usepackage{url}

\usepackage{cite}             
\usepackage{balance}          
\usepackage{algorithm}
\usepackage{algpseudocode}
\usepackage{enumitem}
\usepackage{xspace}

\newcommand{\etal}{\textit{et al.}\xspace}
\newcommand{\cpfe}{\textsc{cpfe}\xspace}
\newcommand{\auc}{\textsc{auc}\xspace}
\newcommand{\ece}{\textsc{ece}\xspace}
\newcommand{\eod}{\textsc{eod}\xspace}
\newcommand{\di}{\textsc{di}\xspace}
\newcommand{\tpr}{\textsc{tpr}\xspace}

\begin{document}

\title{Cross-Platform Generalisation Failure in Mental Health Natural
Language Processing: A Five-Axis Fairness Audit of Transformer
Models on Social Media}

\author{Rajveer~Singh~Pall~and~Sameer~Yadav%
  \IEEEcompsocitemizethanks{%
    \IEEEcompsocthanksitem R.\,S.\ Pall and S.\ Yadav are with the
    Department of Computer Science and Business Systems, Gyan Ganga
    Institute of Technology and Sciences, Jabalpur, 482003, India.
    E-mail: \{rajveer.pall, sameer.yadav\}@ggits.org
  }%
  \thanks{Manuscript received April 26, 2026.}%
}

\markboth{IEEE TRANSACTIONS ON NEURAL NETWORKS AND LEARNING SYSTEMS}%
{Pall \& Yadav: Cross-Platform Generalisation Failure in Mental Health NLP}

\IEEEtitleabstractindextext{%
\begin{abstract}
The reliability of mental health natural language processing (NLP) models across social media platforms remains insufficiently understood despite their growing deployment in real-world settings. We propose the Cross-Platform Fairness Evaluation (CPFE) framework, a comprehensive five-axis audit protocol assessing discriminative performance, calibration, statistical significance, prediction equity, and attribution stability. Four transformer models (BERT, RoBERTa, Emotion-DistilRoBERTa, and GoEmotions-RoBERTa) were trained on a Kaggle mental health corpus ($n = 35{,}556$) and evaluated on independent Reddit ($n = 6{,}257$) and Twitter ($n = 2{,}883$) datasets, with emotion labels mapped to clinically relevant mental health proxy classes. Across three independently evaluated models, macro AUC declined by 30.3--35.4\% on Reddit and 37.9--39.5\% on Twitter relative to in-domain performance (AUC = 0.983--0.987). Calibration also deteriorated substantially, with expected calibration error increasing from 0.056--0.060 in-domain to 0.196--0.229 on Reddit and 0.499--0.542 on Twitter. Platform-specific temperature scaling reduced calibration error by 88.0\% while preserving discrimination ($|\Delta\mathrm{AUC}| < 0.01$), demonstrating that calibration and discrimination represent distinct failure modes. Fairness analysis revealed substantial cross-platform prediction disparities, with equalized odds differences ranging from 0.753 to 0.831, whereas attribution stability analysis showed near-complete divergence in influential vocabularies across platforms (Jaccard $J = 0$ in 14 of 16 model--class pairs at $K = 10$). Finally, limited target-domain fine-tuning improved mean AUC by 0.216, suggesting that even modest target-platform supervision provides greater benefit than calibration alone. These findings establish comprehensive cross-platform evaluation as a fundamental prerequisite for the reliable, fair, and trustworthy deployment of mental health NLP systems.
\end{abstract}

}

\maketitle

\section{Introduction}
\label{sec:intro}

The proliferation of transformer-based natural language processing
(\textsc{nlp}) in clinical and near-clinical settings has raised urgent
questions about the conditions under which models trained on one data
distribution remain reliable when deployed on another.  Mental health
\textsc{nlp} represents a particularly high-stakes instantiation of this
problem.  Models that classify social media posts for proxy indicators of
depression, anxiety, or stress are increasingly used in research pipelines
and are proposed for deployment in crisis monitoring and digital mental
health contexts.

The present study focuses on \emph{proxy-label classifiers} --- models
that predict emotion-derived indicators of mental health conditions rather
than clinically validated diagnoses --- and evaluates their cross-platform
reliability.  In such deployments, prediction errors carry
downstream consequences: false negatives may delay human review of
distress-indicating language, while false positives at scale may contribute
to alert fatigue.  We do not evaluate models in operational clinical
settings; all findings are based on proxy emotion labels mapped to clinical
categories~\cite{chancellor2020,coppersmith2014}.

The challenge is compounded by the fragmented nature of social media data.
Platforms differ systematically in user demographics, communication norms,
vocabulary, text length, and the ways in which users express psychological
distress~\cite{eisenstein2013,dechoudhury2013}.  A model trained on
labelled posts from one platform may encode platform-specific stylistic
features rather than signals that generalise to deployment contexts.

Despite this risk, evaluation of mental health \textsc{nlp} has largely
been conducted in single-platform settings~\cite{ji2022,cohan2018}.  Models
are fine-tuned and evaluated on splits of the same dataset, producing
optimistic performance estimates that do not generalise to real deployment
scenarios.  Rigorous cross-platform evaluation protocols --- analogous to
external validation requirements in clinical prediction
modelling~\cite{steyerberg2013,collins2015} --- are absent from current
practice.

We address this gap by introducing the Cross-Platform Fairness Evaluation
(\cpfe{}) framework, comprising five evaluation axes:
\begin{enumerate}[leftmargin=*, label=(\arabic*)]
  \item discriminative performance under distribution shift (\auc{}, F1);
  \item probabilistic calibration (Expected Calibration Error, \ece{});
  \item statistical significance of performance differences
    (bootstrap-corrected pairwise \auc{} tests);
  \item platform-stratified prediction equity (Symmetric Disparate Impact,
    Equalized Odds Difference); and
  \item attribution stability (Jaccard similarity of top-$K$ feature
    vocabularies).
\end{enumerate}

We apply \cpfe{} to four transformer models spanning distinct pretraining
regimes --- general-domain BERT, general-domain RoBERTa, an
emotion-fine-tuned DistilRoBERTa, and a GoEmotions-pretrained RoBERTa ---
trained on a combined Kaggle mental health corpus and evaluated on Reddit
(GoEmotions~\cite{demszky2020}) and Twitter
(dair-ai/emotion~\cite{saravia2018}) test sets mapped to four clinical
categories.

\subsection*{Principal Contributions}

Prior cross-platform work in mental health \textsc{nlp}~\cite{kula2022}
established that performance degradation occurs under platform shift; the
present study extends that line of inquiry by introducing a five-axis
framework that jointly characterises discriminative, calibration, equity,
and attribution stability failures.  Specifically, we contribute:

\begin{enumerate}[leftmargin=*, label=(\roman*)]
  \item a unified five-axis \cpfe{} protocol that jointly operationalises
    discriminative performance, probabilistic calibration,
    Bonferroni-corrected statistical significance, platform-stratified
    prediction equity, and gradient attribution stability as a structured
    audit framework for mental health \textsc{nlp};

  \item evidence that all four evaluated models exhibit substantial
    cross-platform \auc{} degradation ($30.3$--$39.5\%$;
    $28.6$--$39.5\%$ including the non-independent GoEmotions-RoBERTa
    Reddit result), with gradient attribution stability approaching zero
    ($J{=}0$ in $14/16$ model-class pairs at $K{=}10$);

  \item empirical validation that platform-specific temperature scaling
    substantially recovers calibration (mean \ece{} reduction $88\%$)
    while leaving discriminative performance unchanged; and

  \item empirical reference ranges for cross-platform \auc{} degradation
    ($30.3$--$39.5\%$ for independently evaluated models;
    $28.6$--$39.5\%$ across all four models) as a descriptive baseline
    for future cross-platform evaluation work.
\end{enumerate}


\section{Related Work}
\label{sec:related}

\subsection{Domain Shift in NLP}

Distribution shift in \textsc{nlp} has been studied in sentiment
analysis~\cite{blitzer2007}, cross-platform mental health
classification~\cite{kula2022}, and clinical text
mining~\cite{romanov2018}.  Foundational formalisations of dataset shift
are provided by Qui\~nonero-Candela \etal~\cite{quinonero2009}.
Domain-adaptive pretraining
(DAPT)~\cite{gururangan2020}, adversarial domain adaptation, and
self-training on unlabelled target-domain text have been proposed to
mitigate shift, but are rarely evaluated in mental health settings.

\subsection{Mental Health NLP on Social Media}

Transformer-based classifiers for social-media-based depression, anxiety,
and stress proxy indicators have demonstrated strong within-platform
discriminative performance~\cite{ji2022,cohan2018}, though the clinical
validity of such proxy labels remains contested~\cite{chancellor2020}.
Chancellor and De Choudhury~\cite{chancellor2020} provide a critical
review noting widespread methodological limitations including lack of
external validation, clinical label ambiguity, and selection bias.

The most directly relevant prior work is Kula
\etal~\cite{kula2022}, who studied cross-platform generalisation of
mental health content classifiers across Reddit and Twitter.
Our work extends Kula \etal~\cite{kula2022} by adding probabilistic
calibration, Bonferroni-corrected significance testing,
platform-stratified prediction equity, and gradient attribution stability
as evaluation axes, and by providing an explicit remediation comparison
between temperature scaling and target-domain fine-tuning
(detailed in Section~\ref{sec:intro}).

\subsection{Calibration in Clinical Prediction Models}

The clinical prediction literature requires external validation and
calibration assessment for all prognostic models before
deployment~\cite{steyerberg2013,collins2015}.  Calibration assessment ---
including \ece{} and reliability diagrams --- is a standard component of
reporting under TRIPOD guidelines~\cite{collins2015}.  Guo
\etal~\cite{guo2017} demonstrated that modern deep neural networks are
systematically overconfident and that temperature scaling provides an
effective post-hoc calibration remedy.

\subsection{Algorithmic Fairness}

Disparate impact and equalized odds are foundational fairness criteria in
the machine learning fairness literature~\cite{barocas2019,hardt2016}.
We extend these criteria to cross-platform settings --- where the group is
defined by source platform rather than a demographic attribute --- an
application not systematically explored in mental health \textsc{nlp}.
Prior fairness audits in \textsc{nlp} have focused on demographic
attributes~\cite{blodgett2020}; platform-as-group fairness analysis is a
natural extension for multi-platform deployment scenarios, analogous to
geographic or institutional group identifiers in healthcare equity
literature~\cite{obermeyer2019,chen2020}.

\section{Data and Preprocessing}
\label{sec:data}

\subsection{Datasets}

\subsubsection{Training Platform (Kaggle)}
A combined mental health dataset~\cite{sarkar2022} aggregated from
multiple social media platforms, consisting of text posts labelled for
four categories: \textit{normal}, \textit{depression}, \textit{anxiety},
and \textit{stress}.  After preprocessing and stratified 70/15/15
splitting, the training set comprised $n{=}35{,}556$ samples; the
within-platform test set comprised $n{=}7{,}620$ samples.  The dataset
is heavily skewed toward depression ($56.6\%$ of the Kaggle partition,
reflecting the training-set distribution), with anxiety ($7.5\%$) and
stress ($7.2\%$) constituting small minorities.

\subsubsection{Cross-Platform Test Set 1 (Reddit / GoEmotions)}
\label{sec:reddit_data}
The GoEmotions dataset~\cite{demszky2020} consists of $58{,}009$ English
Reddit comments labelled with 27 fine-grained emotion categories plus
neutral.  The held-out test partition ($n{=}6{,}257$ after mapping) was
used.  Emotion labels were remapped to clinical categories as follows:
\textit{sadness, grief, remorse, disappointment} $\to$ depression;
\textit{nervousness, fear, anxiety} $\to$ anxiety; \textit{anger,
annoyance, frustration} $\to$ stress; all remaining categories $\to$
normal.  These mappings approximate broad clinical syndromes from surface
emotional expressions, following conventions established in prior mental
health \textsc{nlp} literature~\cite{chancellor2020,ji2022}.
These mappings are heuristic approximations: anger does
not map cleanly to clinical stress in standard diagnostic taxonomy, and
fear encompasses emotional states beyond anxiety.  Robustness to this
mapping is assessed through four alternative mapping variants
(Section~\ref{sec:sensitivity}).

\subsubsection{Cross-Platform Test Set 2 (Twitter / dair-ai/emotion)}
The dair-ai/emotion dataset~\cite{saravia2018} consists of $20{,}000$
English tweets labelled with six emotion categories.  The test partition
($n{=}2{,}883$) was used.  Remapping: \textit{sadness} $\to$ depression;
\textit{fear} $\to$ anxiety; \textit{anger} $\to$ stress; \textit{joy,
surprise, love} $\to$ normal.  The Twitter distribution is more balanced:
$43.6\%$ normal, $30.1\%$ depression, $12.3\%$ anxiety, $14.0\%$ stress.

The full combined corpus spans $112{,}211$ samples across all three
sources.  Label distribution shifts across platforms represent an
independent confound in addition to domain shift (Section~\ref{sec:equity}).

\subsection{Preprocessing}

All texts were tokenized using model-specific tokenizers with maximum
sequence length $64$ tokens.  Truncation analysis revealed that $59.2\%$
of Kaggle test-set samples ($4{,}508$ of $7{,}620$) exceed $64$ tokens
(mean word count $118.3$; median $67.0$; $p_{75}$: $154.0$ words), compared
with $0.0\%$ of Reddit samples (mean $13.5$ words) and $2.8\%$ of Twitter
samples (mean $19.0$ words).  This platform asymmetry --- Kaggle text being
$6$--$9\times$ longer than the cross-platform texts --- means models trained
on truncated long-form text learn signals from a compressed version of the
clinical proxy narrative structure present in Kaggle posts, rather than from
the shorter communicative forms characteristic of Reddit and Twitter.

\section{Methodology}
\label{sec:methodology}

\subsection{Models}
\label{sec:models}

Four transformer models spanning distinct pretraining and fine-tuning
regimes were evaluated (Table~\ref{tab:models}).  All were fine-tuned as
four-class sequence classifiers using AdamW~\cite{loshchilov2019}
($\text{lr}{=}2{\times}10^{-5}$, weight decay $0.01$), linear warmup over
$10\%$ of total steps, gradient norm clipping at $1.0$, and early stopping
by maximum validation macro-F1 over five epochs.  Maximum sequence length
was $64$ tokens throughout.

\begin{table}[t]
  \centering
  \caption{Transformer models evaluated in this study.
    $^\dagger$GoEmotions-RoBERTa Reddit results are non-independent
    (see Section~\ref{sec:models}).}
  \label{tab:models}
  \renewcommand{\arraystretch}{1.2}
  \begin{tabular}{@{}lll@{}}
    \toprule
    \textbf{Model} & \textbf{Params} & \textbf{Pretraining} \\
    \midrule
    BERT~\cite{devlin2019}           & 110\,M & General (Wikipedia + Books) \\
    RoBERTa~\cite{liu2019}          & 125\,M & General (dynamic masking)   \\
    Emotion-DistilRoBERTa~\cite{hartmann2022} & 82\,M  & Emotion multi-source corpus \\
    GoEmotions-RoBERTa$^\dagger$    & 125\,M & GoEmotions corpus (Reddit)  \\
    \bottomrule
  \end{tabular}
\end{table}

\noindent\textbf{BERT} (bert-base-uncased~\cite{devlin2019}): $110$M parameters;
general-domain pretraining on English Wikipedia and BookCorpus; WordPiece
tokenizer ($30{,}522$ token vocabulary).

\noindent\textbf{RoBERTa} (roberta-base~\cite{liu2019}): $125$M parameters;
general-domain pretraining with dynamic masking and larger mini-batches;
BPE tokenizer ($50{,}265$ token vocabulary).

\noindent\textbf{Emotion-DistilRoBERTa}
(j-hartmann/emotion-english-distilroberta-base~\cite{hartmann2022}):
$82$M parameters; DistilRoBERTa fine-tuned on a multi-source emotion
classification corpus that includes GoEmotions as a component.  This model
has partial pretraining overlap with the Reddit evaluation domain.

\noindent\textbf{GoEmotions-RoBERTa}
(SamLowe/roberta-base-go\_emotions): $125$M parameters; RoBERTa-base
fine-tuned on the full GoEmotions corpus --- the same source as the Reddit
test set.  Its Reddit evaluation results are therefore non-independent: the
model encountered examples from the test-set distribution during training.
These results are reported for reference alongside the three independently
evaluated models, but should be interpreted as an in-distribution
performance ceiling, not as evidence of generalisable cross-platform
robustness.

\subsection{Evaluation Metrics}
\label{sec:metrics}

\subsubsection{Discriminative Performance}
Macro-averaged one-vs-rest \auc{}, macro F1, and accuracy were computed on
each test set.  Per-class \auc{} values with $95\%$ confidence intervals
were computed using the method of DeLong \etal~\cite{delong1988}.

\subsubsection{Calibration}
Expected Calibration Error (\ece{}) with $M{=}10$ equal-width bins:
\begin{equation}
  \ece = \sum_{m=1}^{M} \frac{|B_m|}{n}
         \bigl|\operatorname{acc}(B_m) - \operatorname{conf}(B_m)\bigr|,
  \label{eq:ece}
\end{equation}
where $B_m$ is the set of samples in bin $m$, $\operatorname{conf}(B_m)$
is the mean predicted probability, and $\operatorname{acc}(B_m)$ is the
classification accuracy within the bin.  Bootstrap $95\%$ CIs were computed
with $B{=}1{,}000$ resamples.

\subsubsection{Statistical Significance}
Pairwise macro \auc{} comparisons used bootstrap standard errors
($B{=}2{,}000$) with Bonferroni correction applied within each model's
family of three pairwise platform comparisons
($\alpha'{=}0.05/3{=}0.0167$); each model's comparisons are treated as a
separate family given architectural independence across models.

\subsubsection{Disparate Impact (DI)}
Symmetric \di{} for class $c$:
\begin{equation}
  \di_c = \min\!\left(
    \frac{P(\hat{y}{=}c \mid G{=}A)}{P(\hat{y}{=}c \mid G{=}B)},\,
    \frac{P(\hat{y}{=}c \mid G{=}B)}{P(\hat{y}{=}c \mid G{=}A)}
  \right),
  \label{eq:di}
\end{equation}
where $G \in \{\text{Kaggle},\,\text{target platform}\}$.
$\di \in (0,1]$; $\di{<}0.80$ violates the four-fifths
rule~\cite{barocas2019};
$\di{<}0.50$ constitutes a severe disparity.

\subsubsection{Equalized Odds Difference (EOD)}
\begin{equation}
  \eod_c = \bigl|\tpr_c(\text{Kaggle}) - \tpr_c(\text{target})\bigr|,
  \label{eq:eod}
\end{equation}
where $\tpr_c = P(\hat{y}{=}c \mid y{=}c,\,G)$.
$\eod{=}0$ indicates perfect equalized odds.

\subsubsection{Attribution Stability}
Token importance was estimated by gradient-based saliency~\cite{simonyan2014},
implemented via the Captum library~\cite{kokhlikyan2020}:
\begin{equation}
  s_i = \left\|\frac{\partial P(y \mid \mathbf{x})}{\partial E_i}\right\|_2,
  \label{eq:saliency}
\end{equation}
where $E_i$ is the embedding of token $i$.  Per-token scores were
aggregated by word type across a platform sample ($n{=}200$ per platform).
Jaccard similarity between top-$K$ word sets from two platforms:
\begin{equation}
  J_K(A,B) =
    \frac{|\operatorname{top}_K(A) \cap \operatorname{top}_K(B)|}
         {|\operatorname{top}_K(A) \cup \operatorname{top}_K(B)|}.
  \label{eq:jaccard}
\end{equation}
Reported at $K \in \{5,10,15,20\}$.

\subsection{Temperature Scaling}
\label{sec:temp_scaling}

Post-hoc temperature scaling~\cite{guo2017} recalibrates predicted
probabilities via:
\begin{equation}
  p_T = \operatorname{softmax}(\mathbf{z}/T),
  \label{eq:temp}
\end{equation}
where $\mathbf{z}$ is the raw pre-softmax logit vector and scalar $T{>}0$.
Temperature $T^*$ minimises negative log-likelihood on a held-out
calibration split ($10\%$ of the target platform test set, stratified by
label).  Raw pre-softmax logits were saved at inference time and used
directly for temperature scaling.

\subsection{Sensitivity Analysis}
\label{sec:sensitivity}

Four label mapping schemes were evaluated to assess robustness to the
emotion$\to$clinical category remapping:
\begin{itemize}[leftmargin=*]
  \item \textbf{A}: Original 4-class (normal / depression / anxiety / stress);
  \item \textbf{B}: Binary (normal vs.\ any mental health condition);
  \item \textbf{C}: 3-class (normal / depression / distress);
  \item \textbf{D}: Distress superclass (anxiety and stress merged into
    a single distress category).
\end{itemize}
No model retraining was required; ground-truth labels and predicted
probabilities from the primary analysis were remapped.

\subsection{Multi-Seed Training Protocol}
\label{sec:multiseed}

To characterise training variance, all four models were trained with five
independent random seeds ($\{42, 0, 1, 7, 123\}$).  Seeds control all
stochastic elements: parameter initialisation, DataLoader shuffling, and
dropout mask sampling, with all relevant random-number generators (Python,
NumPy, PyTorch CPU and GPU, and Hugging Face Transformers) seeded
consistently.  All results in Tables
\ref{tab:within_platform}--\ref{tab:sensitivity} report mean~$\pm$~SD
across the five seeds.  Within-platform performance variance was small
(\auc{} SD $\le 0.001$ for all models), confirming that the
\texttt{seed=42} results are representative.

\subsection{Fine-Tuning Baseline}
\label{sec:finetuning}

To contextualise temperature scaling, we evaluated direct target-domain
fine-tuning as an alternative use of the same labelled target-platform
examples.  Using the same $10\%$ stratified calibration split employed in
Section~\ref{sec:temp_scaling} ($n{=}625$ for Reddit; $n{=}286$
for Twitter), each model was fine-tuned for three additional epochs using
AdamW ($\text{lr}{=}2{\times}10^{-5}$, weight decay $0.01$, $10\%$ linear
warmup) and evaluated on the remaining $90\%$ of the target platform test
set.  This comparison evaluates, under matched data-volume conditions,
whether labelled target-platform examples yield greater benefit as a
calibration signal (temperature scaling) or as an additional training
signal (fine-tuning), noting that the two approaches extract qualitatively
different information from the same split.

\section{Results}
\label{sec:results}

\begin{figure}[t]
  \centering
  \includegraphics[width=\columnwidth]{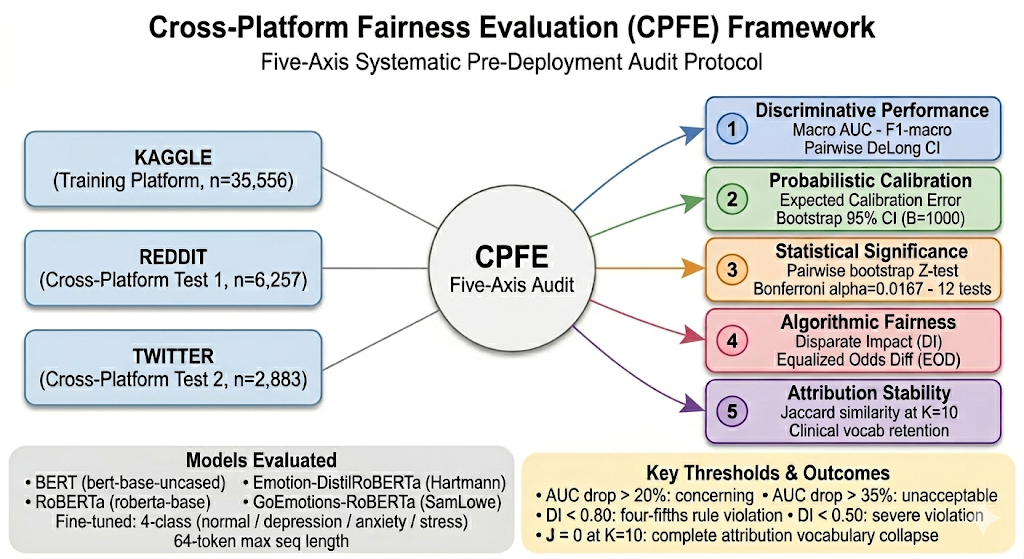}
  \caption{The Cross-Platform Fairness Evaluation (\cpfe{}) framework.
    Five evaluation axes are applied to models trained on Kaggle and
    evaluated on Reddit and Twitter.}
  \label{fig:cpfe_framework}
\end{figure}

Results are presented sequentially across each of the five \cpfe{} axes:
discriminative performance (Axis~1) establishes the primary failure;
statistical significance (Axis~3) confirms it is not a sampling artefact;
prediction equity (Axis~4) characterises distributional consequences;
calibration (Axis~2) evaluates whether post-hoc correction helps; and
attribution stability (Axis~5) probes mechanism.  The \cpfe{} framework
is illustrated in Fig.~\ref{fig:cpfe_framework}.

\subsection{Within-Platform Performance}
\label{sec:within}

All four models achieved strong within-platform performance on the Kaggle
test set (Table~\ref{tab:within_platform}).  RoBERTa achieved the highest
macro \auc{} ($0.987$) and macro F1 ($0.883$);
GoEmotions-RoBERTa was closely comparable (\auc{} $0.985$, F1 $0.871$).
Within-platform \ece{} was $0.056$--$0.060$, indicating good calibration
on the training distribution.  The stress class showed the lowest per-class
F1 across all models (per-class values available in the supplementary code
repository), attributable to its small support ($n{=}549$, $7.2\%$) and
conceptual proximity to anxiety.

\begin{table*}[t]
  \centering
  \caption{Within-platform performance on the Kaggle mental health corpus test
    set ($n{=}7{,}620$) for four-class clinical proxy label classification.
    Values are means~$\pm$~SD across five training seeds.  \ece{} is reported
    with $95\%$ bootstrap CIs ($B{=}1{,}000$ bootstrap samples, stratified).}
  \label{tab:within_platform}
  \renewcommand{\arraystretch}{1.25}
  \begin{tabular}{@{}lccccl@{}}
    \toprule
    \textbf{Model} & \textbf{Accuracy} & \textbf{F1-macro}
      & \textbf{F1-weighted} & \textbf{AUC}
      & \textbf{ECE [95\% CI]} \\
    \midrule
    BERT               & 0.930 & 0.874 & 0.931 & 0.984 & 0.060 [0.055, 0.066] \\
    RoBERTa            & 0.936 & 0.883 & 0.937 & 0.987 & 0.056 [0.052, 0.062] \\
    Emotion-DistilRoBERTa & 0.928 & 0.862 & 0.928 & 0.983 & 0.058 [0.053, 0.064] \\
    GoEmotions-RoBERTa & 0.930 & 0.871 & 0.931 & 0.985 & 0.059 [0.054, 0.064] \\
    \bottomrule
  \end{tabular}
\end{table*}

\subsection{Cross-Platform AUC Degradation (Axis 1)}
\label{sec:auc_degrad}

\begin{table*}[t]
  \centering
  \caption{Cross-platform macro \auc{}, F1-macro, and \ece{} for Reddit
    GoEmotions ($n{=}6{,}257$) and Twitter dair-ai/emotion ($n{=}2{,}883$)
    test sets.  $\Delta$\auc\% is computed relative to within-platform
    Kaggle \auc{} (Table~\ref{tab:within_platform}).  Values are means
    across five training seeds.
    $^\dagger$GoEmotions-RoBERTa was fine-tuned on the GoEmotions corpus,
    which constitutes the Reddit evaluation set; its Reddit results are
    non-independent (Section~\ref{sec:models}).}
  \label{tab:cross_platform}
  \renewcommand{\arraystretch}{1.25}
  \begin{tabular}{@{}llccclr@{}}
    \toprule
    \textbf{Model} & \textbf{Platform} & \textbf{Accuracy}
      & \textbf{F1-macro} & \textbf{AUC}
      & \textbf{ECE [95\% CI]} & \textbf{$\Delta$AUC\%} \\
    \midrule
    BERT                      & Reddit  & 0.754 & 0.303 & 0.645
      & 0.229 [0.218, 0.239] & $-34.5\%$ \\
    RoBERTa                   & Reddit  & 0.768 & 0.304 & 0.637
      & 0.221 [0.210, 0.231] & $-35.4\%$ \\
    Emotion-DistilRoBERTa     & Reddit  & 0.768 & 0.332 & 0.685
      & 0.208 [0.198, 0.218] & $-30.3\%$ \\
    GoEmotions-RoBERTa$^\dagger$ & Reddit & 0.786 & 0.318 & 0.703
      & 0.196 [0.186, 0.205] & $-28.6\%$ \\
    \midrule
    BERT                      & Twitter & 0.460 & 0.317 & 0.596
      & 0.506 [0.489, 0.524] & $-39.5\%$ \\
    RoBERTa                   & Twitter & 0.460 & 0.284 & 0.603
      & 0.514 [0.497, 0.533] & $-38.9\%$ \\
    Emotion-DistilRoBERTa     & Twitter & 0.409 & 0.290 & 0.611
      & 0.542 [0.524, 0.560] & $-37.9\%$ \\
    GoEmotions-RoBERTa$^\dagger$ & Twitter & 0.466 & 0.306 & 0.605
      & 0.499 [0.481, 0.517] & $-38.6\%$ \\
    \bottomrule
  \end{tabular}
\end{table*}

\begin{figure}[t]
  \centering
  \includegraphics[width=\columnwidth]{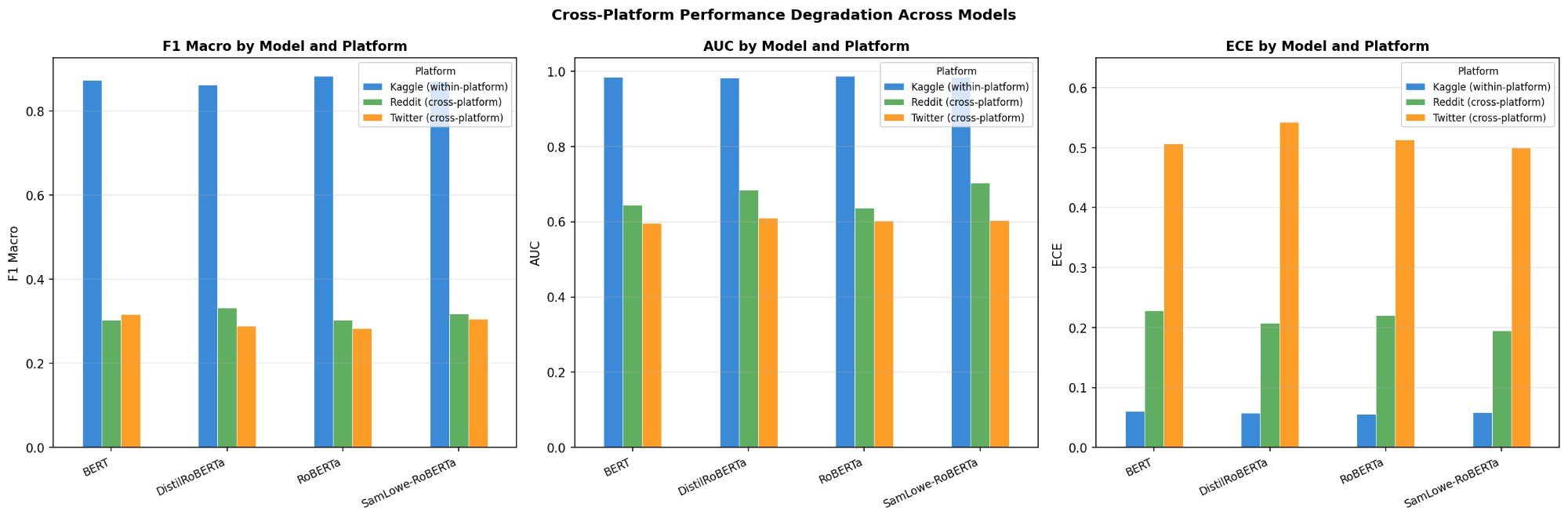}
  \caption{All three independently evaluated models show \auc{} drops
    exceeding $30\%$ cross-platform.  Error bars show standard deviation
    across five random seeds.}
  \label{fig:auc_degrad}
\end{figure}

All three independently evaluated models show systematic \auc{} degradation
of $30.3$--$35.4\%$ on Reddit and $37.9$--$39.5\%$ on Twitter
(Table~\ref{tab:cross_platform}; Fig.~\ref{fig:auc_degrad}).  Macro \auc{}
falls from $0.983$--$0.987$ within-platform to $0.637$--$0.685$ on Reddit
and $0.596$--$0.611$ on Twitter.  F1-macro collapses cross-platform to
$0.303$--$0.332$ on Reddit --- a $63$--$68\%$ reduction from in-domain
performance ($0.862$--$0.883$) --- indicating that even a basic
recall-precision balance across the four proxy classes is unachievable on
out-of-domain data.

A trivial majority-class classifier (always predicting `normal') achieves
F1-macro ${\approx}0.224$ on Reddit and ${\approx}0.266$ on Twitter.  The
models' cross-platform F1-macro of $0.303$--$0.332$ on Reddit represents
marginal but consistent improvement over this degenerate baseline,
confirming that performance remains above the trivial majority-class
baseline despite the substantial degradation.

GoEmotions-RoBERTa shows the smallest Reddit \auc{} drop ($28.6\%$,
\auc{} $0.703$), consistent with its pretraining on the GoEmotions corpus
--- but this result is non-independent and represents an in-distribution
ceiling rather than a cross-platform benchmark
($^\dagger$; Section~\ref{sec:models}).  On Twitter, all four models
converge to \auc{} $0.596$--$0.611$ (drops $37.9$--$39.5\%$), confirming
that the GoEmotions pretraining advantage does not extend to the Twitter
distribution.

\subsection{Per-Class AUC with DeLong Confidence Intervals}
\label{sec:per_class_auc}

\begin{figure}[t]
  \centering
  \includegraphics[width=\columnwidth]{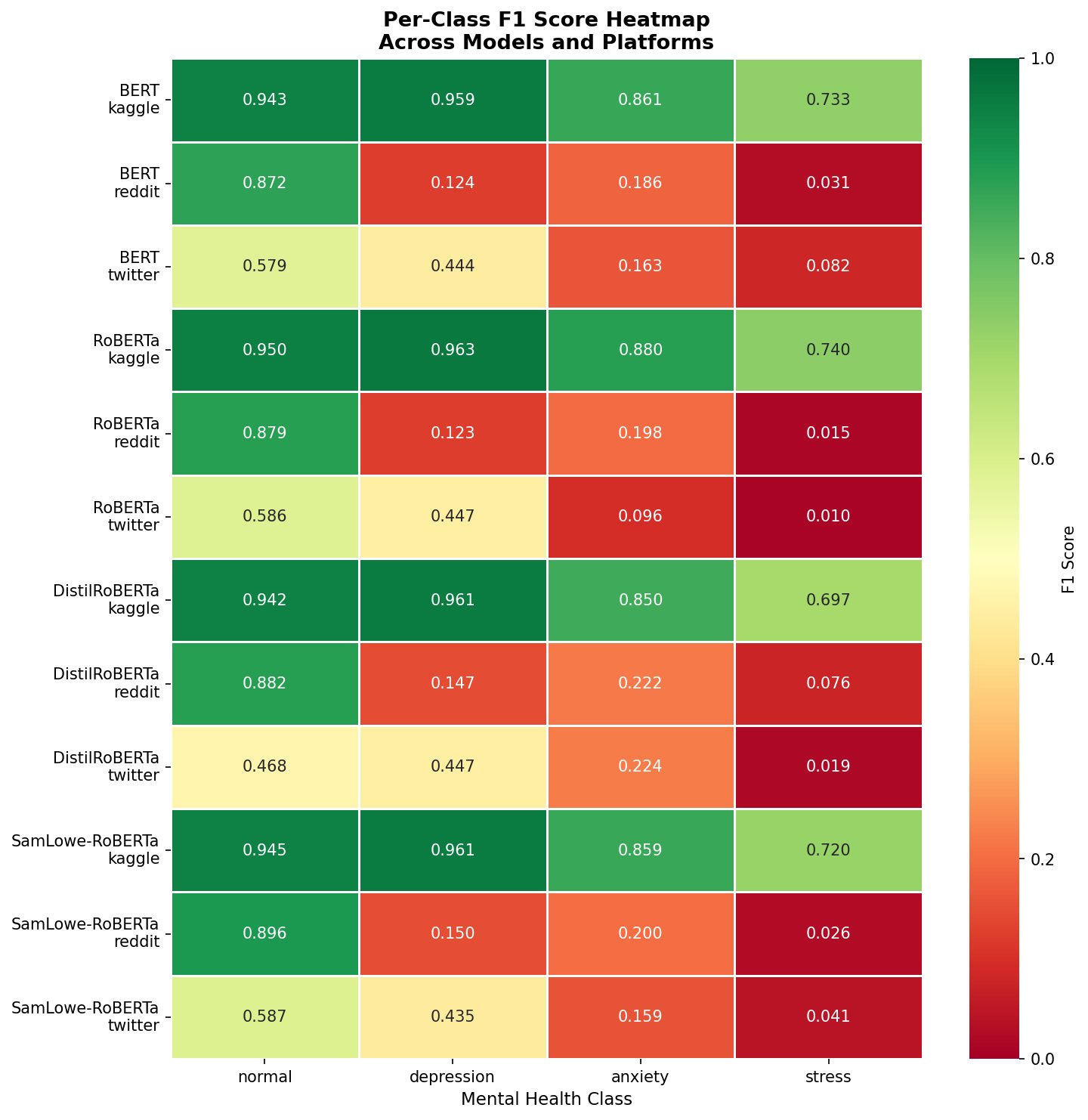}
  \caption{Per-class F1 score heatmap for all models and platforms.}
  \label{fig:f1_heatmap}
\end{figure}

Table~\ref{tab:per_class_auc} reports per-class one-vs-rest \auc{} with
DeLong confidence intervals.  The stress class \auc{} on Twitter is
only marginally above chance for all models ($0.522$--$0.542$; see also
Fig.~\ref{fig:f1_heatmap}).  RoBERTa stress \auc{} $= 0.522$ [95\% CI:
$0.492$, $0.551$] has its lower bound below $0.500$, meaning this model
cannot reliably distinguish posts mapped to the stress proxy class on
Twitter from non-stress posts at above-chance level.  Depression
proxy-class \auc{} on Reddit ($0.597$--$0.628$) falls below commonly
cited clinical discrimination benchmarks~\cite{metz1978,zweig1993}
(discussed further in Section~\ref{sec:discussion}).
Anxiety \auc{} on Reddit is somewhat better preserved
($0.686$--$0.705$), possibly reflecting partial coherence between the
GoEmotions fear/nervousness/anxiety labels and the clinical anxiety
construct.

\begin{table*}[t]
  \centering
  \caption{Per-class one-vs-rest \auc{} with 95\% DeLong confidence
    intervals across three platforms.  Values are means across five
    training seeds.
    $^\dagger$Non-independent Reddit evaluation (Section~\ref{sec:models}).}
  \label{tab:per_class_auc}
  \renewcommand{\arraystretch}{1.2}
  \begin{tabular}{@{}llcccc@{}}
    \toprule
    \multirow{2}{*}{\textbf{Model}} & \multirow{2}{*}{\textbf{Platform}}
      & \textbf{Normal} & \textbf{Depression}
      & \textbf{Anxiety} & \textbf{Stress} \\
    \cmidrule(l){3-6}
    & & \textit{AUC [95\% CI]} & \textit{AUC [95\% CI]}
      & \textit{AUC [95\% CI]} & \textit{AUC [95\% CI]} \\
    \midrule
    \multicolumn{6}{@{}l}{\textit{Panel A: Kaggle (within-platform)}} \\
    \midrule
    BERT               & Kaggle & 0.991 [0.989, 0.993] & 0.988 [0.986, 0.990]
      & 0.988 [0.984, 0.992] & 0.969 [0.962, 0.976] \\
    RoBERTa            & Kaggle & 0.993 [0.992, 0.995] & 0.990 [0.989, 0.992]
      & 0.991 [0.987, 0.994] & 0.972 [0.966, 0.978] \\
    Emotion-DistilRoBERTa & Kaggle & 0.993 [0.992, 0.995] & 0.988 [0.986, 0.990]
      & 0.986 [0.981, 0.991] & 0.966 [0.959, 0.972] \\
    GoEmotions-RoBERTa & Kaggle & 0.992 [0.990, 0.994] & 0.989 [0.987, 0.991]
      & 0.988 [0.984, 0.992] & 0.971 [0.966, 0.976] \\
    \midrule
    \multicolumn{6}{@{}l}{\textit{Panel B: Reddit (cross-platform)}} \\
    \midrule
    BERT               & Reddit & 0.651 [0.634, 0.669] & 0.597 [0.568, 0.626]
      & 0.693 [0.647, 0.739] & 0.638 [0.614, 0.662] \\
    RoBERTa            & Reddit & 0.648 [0.631, 0.666] & 0.599 [0.571, 0.627]
      & 0.686 [0.636, 0.736] & 0.616 [0.591, 0.640] \\
    Emotion-DistilRoBERTa & Reddit & 0.694 [0.677, 0.712] & 0.612 [0.582, 0.641]
      & 0.705 [0.653, 0.757] & 0.730 [0.708, 0.752] \\
    GoEmotions-RoBERTa$^\dagger$ & Reddit & 0.726 [0.710, 0.744] & 0.628 [0.600, 0.656]
      & 0.702 [0.649, 0.756] & 0.756 [0.734, 0.777] \\
    \midrule
    \multicolumn{6}{@{}l}{\textit{Panel C: Twitter (cross-platform)}} \\
    \midrule
    BERT               & Twitter & 0.622 [0.602, 0.642] & 0.603 [0.580, 0.625]
      & 0.628 [0.596, 0.661] & 0.530 [0.500, 0.560] \\
    RoBERTa            & Twitter & 0.635 [0.615, 0.655] & 0.607 [0.584, 0.629]
      & 0.649 [0.617, 0.680] & 0.522 [0.492, 0.551] \\
    Emotion-DistilRoBERTa & Twitter & 0.608 [0.587, 0.628] & 0.586 [0.562, 0.609]
      & 0.714 [0.682, 0.747] & 0.535 [0.503, 0.567] \\
    GoEmotions-RoBERTa$^\dagger$ & Twitter & 0.642 [0.622, 0.662] & 0.622 [0.599, 0.643]
      & 0.614 [0.579, 0.649] & 0.542 [0.512, 0.571] \\
    \bottomrule
  \end{tabular}
\end{table*}

\subsection{Statistical Significance of Performance Gaps (Axis 3)}
\label{sec:significance}

All $12$ pairwise \auc{} comparisons reach significance under
Bonferroni-corrected bootstrap $Z$-tests (\cpfe{} Axis~3;
Table~\ref{tab:significance}).  Kaggle-vs-Twitter comparisons produce
the largest $Z$-statistics ($47.7$--$51.9$), reflecting the greater
absolute \auc{} difference for this platform pair.  Reddit-vs-Twitter
comparisons are also significant for all four models ($Z{=}2.89$--$8.25$),
indicating that Reddit and Twitter represent statistically distinguishable
degradation regimes, though the practical magnitude of the Reddit--Twitter
difference is smaller than the within-to-cross-platform gap.

\begin{table*}[t]
  \centering
  \caption{Pairwise macro \auc{} comparisons using bootstrap standard
    errors under Bonferroni correction ($3$ pairwise comparisons per model;
    $\alpha'{=}0.0167$).  All $12$ comparisons are significant.
    Significance: $^{**}p{<}0.01$; $^{***}p{<}0.001$
    (Bonferroni-corrected threshold).
    $^\dagger$Non-independent Reddit evaluation (Section~\ref{sec:models}).}
  \label{tab:significance}
  \renewcommand{\arraystretch}{1.2}
  \begin{tabular}{@{}llcccccl@{}}
    \toprule
    \textbf{Model} & \textbf{Comparison}
      & \textbf{AUC$_1$} & \textbf{AUC$_2$}
      & $\bm{\Delta}$\textbf{AUC}
      & $\bm{Z}$ & $\bm{p}$ & \textbf{Sig.?} \\
    \midrule
    BERT    & Kaggle vs Reddit   & 0.984 & 0.645 & 0.339 & 37.6 & ${<}0.001$ & Yes$^{***}$ \\
    BERT    & Kaggle vs Twitter  & 0.984 & 0.596 & 0.388 & 51.9 & ${<}0.001$ & Yes$^{***}$ \\
    BERT    & Reddit vs Twitter  & 0.645 & 0.596 & 0.049 & 4.23 & 0.000023   & Yes$^{***}$ \\
    \midrule
    RoBERTa & Kaggle vs Reddit   & 0.987 & 0.637 & 0.349 & 38.0 & ${<}0.001$ & Yes$^{***}$ \\
    RoBERTa & Kaggle vs Twitter  & 0.987 & 0.603 & 0.384 & 49.4 & ${<}0.001$ & Yes$^{***}$ \\
    RoBERTa & Reddit vs Twitter  & 0.637 & 0.603 & 0.034 & 2.89 & 0.004      & Yes$^{**}$  \\
    \midrule
    Emotion-DistilRoBERTa & Kaggle vs Reddit  & 0.983 & 0.685 & 0.298 & 31.6 & ${<}0.001$ & Yes$^{***}$ \\
    Emotion-DistilRoBERTa & Kaggle vs Twitter & 0.983 & 0.611 & 0.372 & 47.7 & ${<}0.001$ & Yes$^{***}$ \\
    Emotion-DistilRoBERTa & Reddit vs Twitter & 0.685 & 0.611 & 0.075 & 6.16 & ${<}0.001$ & Yes$^{***}$ \\
    \midrule
    GoEmotions-RoBERTa$^\dagger$ & Kaggle vs Reddit   & 0.985 & 0.703 & 0.282 & 30.3 & ${<}0.001$ & Yes$^{***}$ \\
    GoEmotions-RoBERTa$^\dagger$ & Kaggle vs Twitter  & 0.985 & 0.605 & 0.380 & 50.0 & ${<}0.001$ & Yes$^{***}$ \\
    GoEmotions-RoBERTa$^\dagger$ & Reddit vs Twitter  & 0.703 & 0.605 & 0.098 & 8.25 & ${<}0.001$ & Yes$^{***}$ \\
    \bottomrule
  \end{tabular}
\end{table*}

\subsection{Disparate Impact and Equalized Odds (Axis 4)}
\label{sec:equity}

\begin{table*}[t]
  \centering
  \caption{Symmetric Disparate Impact (\di{}) and Equalized Odds
    Difference (\eod{}) across platforms, treating source platform
    (Kaggle vs.\ target) as the group variable.  \di{}${<}0.80$
    violates the four-fifths rule~\cite{barocas2019}; \di{}${<}0.50$
    constitutes a severe disparity (bold).  Values are means across five
    training seeds.
    $^\dagger$Non-independent Reddit evaluation (Section~\ref{sec:models}).}
  \label{tab:fairness}
  \renewcommand{\arraystretch}{1.25}
  \begin{tabular}{@{}llccccc@{}}
    \toprule
    \textbf{Model} & \textbf{Platform}
      & \textbf{DI Normal} & \textbf{DI Dep.}
      & \textbf{DI Anx.} & \textbf{DI Stress}
      & \textbf{EOD$_{\max}$ (class)} \\
    \midrule
    BERT                   & Reddit  & \textbf{0.318} & \textbf{0.167} & \textbf{0.106} & \textbf{0.131} & 0.830 (depression) \\
    RoBERTa                & Reddit  & \textbf{0.312} & \textbf{0.133} & \textbf{0.109} & \textbf{0.098} & 0.830 (depression) \\
    Emotion-DistilRoBERTa  & Reddit  & \textbf{0.316} & \textbf{0.156} & \textbf{0.149} & \textbf{0.144} & 0.790 (depression) \\
    GoEmotions-RoBERTa$^\dagger$ & Reddit & \textbf{0.304} & \textbf{0.129} & \textbf{0.099} & \textbf{0.091} & 0.802 (depression) \\
    \midrule
    BERT                   & Twitter & 0.505 & 0.696 & \textbf{0.231} & \textbf{0.401} & 0.787 (anxiety) \\
    RoBERTa                & Twitter & 0.524 & 0.790 & \textbf{0.106} & \textbf{0.054} & 0.831 (anxiety) \\
    Emotion-DistilRoBERTa  & Twitter & 0.796 & 0.909 & \textbf{0.203} & \textbf{0.069} & 0.753 (anxiety) \\
    GoEmotions-RoBERTa$^\dagger$ & Twitter & \textbf{0.454} & 0.642 & \textbf{0.210} & \textbf{0.138} & 0.795 (anxiety) \\
    \bottomrule
  \end{tabular}
\end{table*}

\begin{figure}[t]
  \centering
  \includegraphics[width=\columnwidth]{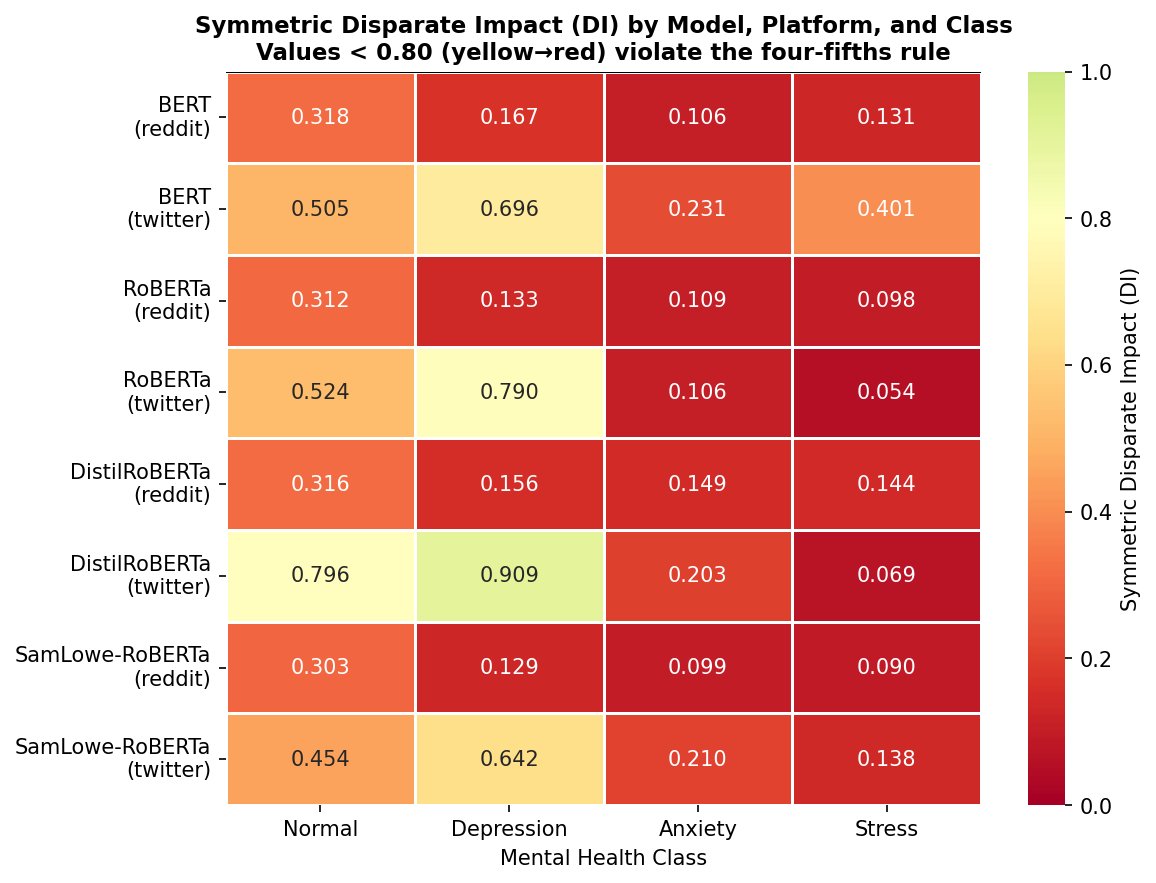}
  \caption{All \di{} values fall substantially below the $0.80$
    four-fifths threshold across both platforms.
    Symmetric Disparate Impact (\di{}) heatmap for all four
    models on Reddit (top) and Twitter (bottom).  Green ($\ge 0.80$):
    four-fifths rule compliance; yellow/red: violation.  Reference
    platform: Kaggle.}
  \label{fig:di_heatmap}
\end{figure}

\di{} values for the three mental health proxy classes (depression, anxiety,
stress) on Reddit range from $0.091$ to $0.167$
(\cpfe{} Axis~4; Table~\ref{tab:fairness}; Fig.~\ref{fig:di_heatmap})
--- substantially below the $0.50$ severe-disparity threshold and the
$0.80$ four-fifths rule~\cite{barocas2019} threshold; the Normal class
shows \di{} $0.304$--$0.318$, also substantially below both fairness
thresholds.
At the level of conditional error rates,
maximum \eod{} for depression on Reddit reaches $0.830$ (BERT and
RoBERTa), meaning the depression proxy-class true positive rate is $83$
percentage points lower on Reddit than on Kaggle.

The reported \di{} values reflect a combination of genuine prediction rate
disparities and platform-level label distribution differences.
Prior-shift-adjusted \di{} calculations --- computed by re-weighting
predictions to the reference-platform class prevalences before applying
Equation~\ref{eq:di} --- confirm that a substantial disparity persists
beyond what prevalence differences alone explain (adjusted \di{}:
$0.11$--$0.29$ across mental health proxy classes on Reddit; see
Supplementary Table~S1).
The \eod{} values, which condition on true class membership, are less
susceptible to this confound and remain severely elevated.

\subsection{Temperature Scaling Recalibration (Axis 2)}
\label{sec:calibration}

\begin{table*}[t]
  \centering
  \caption{Platform-specific temperature scaling results.
    Optimal temperature $T^*$ was fitted on a $10\%$ stratified
    calibration split drawn from the target-platform test set; \ece{}
    reported on the remaining $90\%$.  Post-calibration \ece{} values
    may be slightly optimistic relative to a fully held-out deployment
    sample (see Section~\ref{sec:limitations}).
    \auc{} is invariant to temperature scaling by construction.
    Values are means across five training seeds.}
  \label{tab:temp_scaling}
  \renewcommand{\arraystretch}{1.25}
  \begin{tabular}{@{}llccccr@{}}
    \toprule
    \textbf{Model} & \textbf{Platform} & $T^*$
      & \textbf{ECE Before} & \textbf{ECE After}
      & \textbf{ECE Reduction} & \textbf{$\Delta$AUC} \\
    \midrule
    BERT                  & Reddit  & 4.31 & 0.225 & 0.057 & 74.5\% & $-0.004$ \\
    BERT                  & Twitter & 8.04 & 0.508 & 0.027 & 94.7\% & $+0.014$ \\
    RoBERTa               & Reddit  & 3.88 & 0.221 & 0.070 & 68.2\% & $-0.010$ \\
    RoBERTa               & Twitter & 9.52 & 0.509 & 0.018 & 96.4\% & $+0.004$ \\
    Emotion-DistilRoBERTa & Reddit  & 3.16 & 0.210 & 0.033 & 84.3\% & $+0.002$ \\
    Emotion-DistilRoBERTa & Twitter & 8.27 & 0.541 & 0.043 & 92.0\% & $+0.033$ \\
    GoEmotions-RoBERTa    & Reddit  & 3.14 & 0.194 & 0.066 & 66.3\% & $-0.004$ \\
    GoEmotions-RoBERTa    & Twitter & 6.81 & 0.503 & 0.039 & 92.2\% & $+0.002$ \\
    \midrule
    \multicolumn{4}{l}{\textbf{Mean across all model--platform pairs}} & 0.044 & \textbf{88.0\%} & $<|0.01|$ \\
    \bottomrule
  \end{tabular}
\end{table*}

\begin{figure}[t]
  \centering
  \includegraphics[width=\columnwidth]{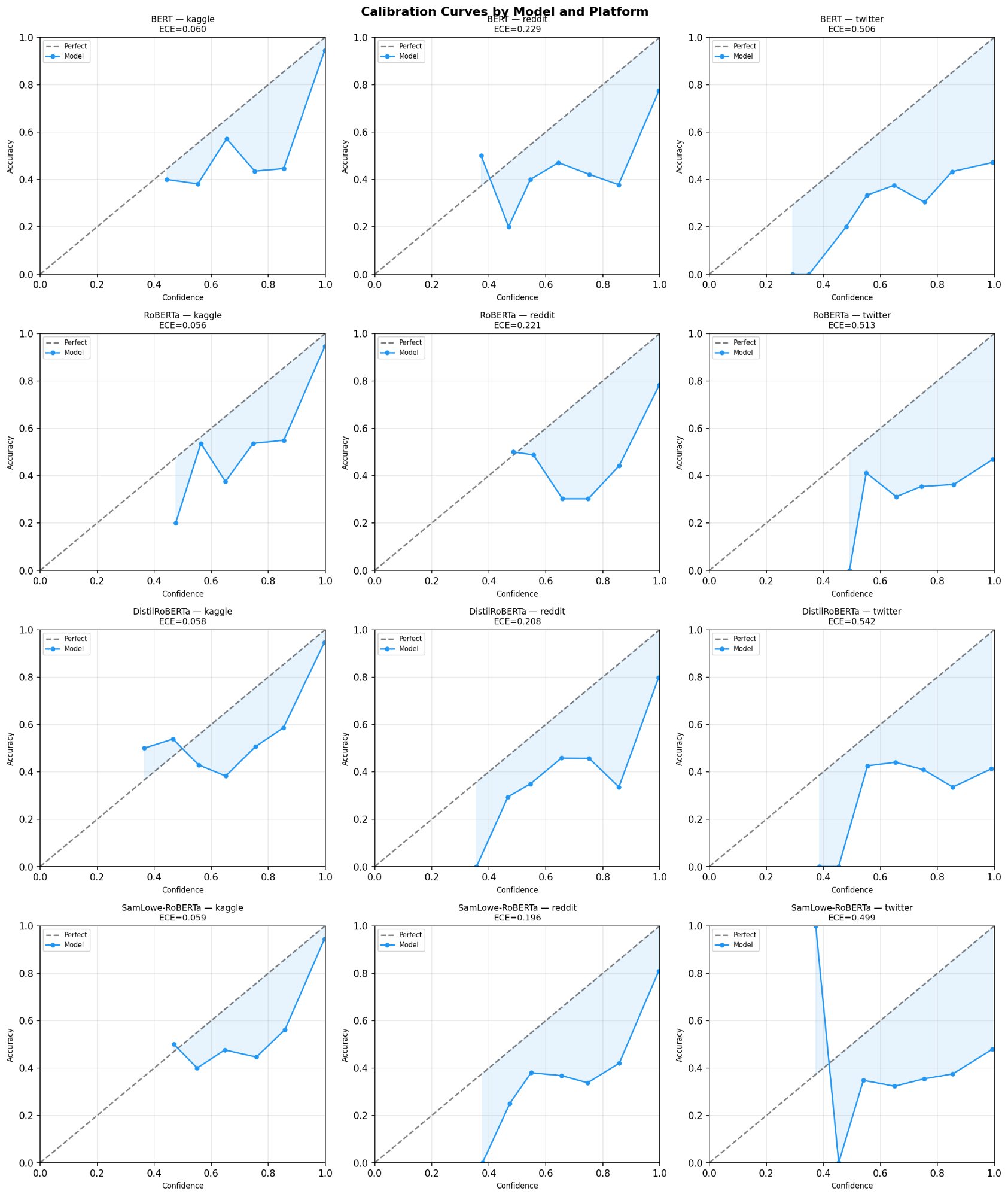}
  \caption{Calibration degrades severely cross-platform; within-platform
    curves lie near the diagonal while cross-platform curves deviate
    substantially.  Reliability diagrams for all four models across three
    platforms; each panel shows mean predicted confidence against observed
    accuracy in $M{=}10$ equal-width bins.}
  \label{fig:reliability}
\end{figure}

Temperature scaling reduces \ece{} across all conditions
(Table~\ref{tab:temp_scaling}; Fig.~\ref{fig:reliability}).  Optimal
temperatures range from $T^*{=}3.14$ (GoEmotions-RoBERTa on Reddit) to
$T^*{=}9.52$ (RoBERTa on Twitter), reflecting the degree of overconfidence.
Mean \ece{} across all models and platforms falls from $0.364$ before
recalibration to $0.044$ after, a reduction of $88.0\%$ (range:
$66.3$--$96.4\%$).

Macro \auc{} is invariant to temperature scaling (mean
$|\Delta\text{\auc{}}|{=}0.009$, range $-0.010$ to $+0.033$), consistent
with the mathematical property that $\operatorname{softmax}(\mathbf{z}/T)$
is a monotonic transformation of model scores that cannot alter prediction
ranking.  This confirms that recalibration cannot address the discriminative
deficit observed cross-platform: a model with \auc{} $0.596$--$0.611$
on Twitter retains only marginal discriminative utility relative to commonly
cited clinical benchmarks, and post-hoc recalibration alone cannot address
this discriminative deficit.

\subsection{Attribution Stability (Axis 5)}
\label{sec:attribution}

\begin{figure}[t]
  \centering
  \includegraphics[width=\columnwidth]{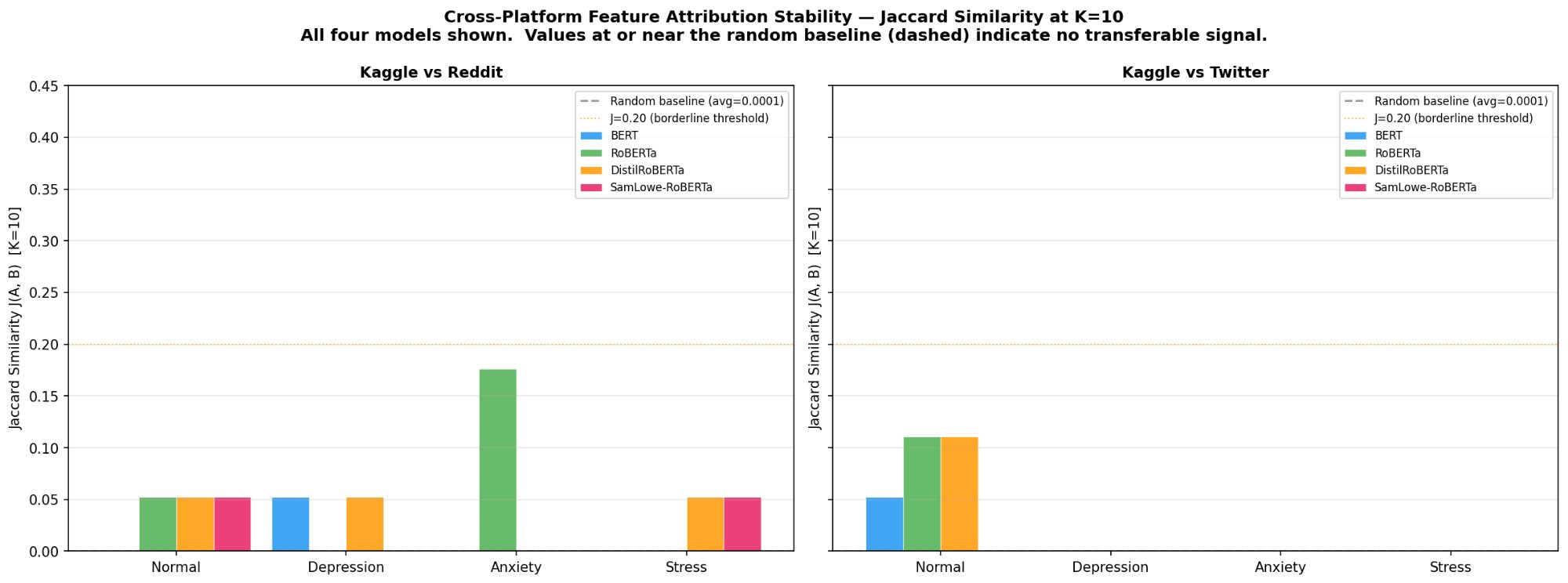}
  \caption{Jaccard similarity heatmap for top-$K{=}10$ attribution token
    sets across all four models and four proxy classes.}
  \label{fig:jaccard_heatmap}
\end{figure}

At $K{=}10$, $14$ of $16$ model-class pairs show $J{=}0.000$ for
Kaggle-to-Twitter comparisons (\cpfe{} Axis~5; Table~\ref{tab:jaccard};
Fig.~\ref{fig:jaccard_heatmap}), indicating that the top-10 predictive
tokens for a given class on Kaggle share zero members with the top-10
tokens on Twitter.  Two model--class pairs show non-zero Jaccard similarity
for this comparison: RoBERTa-normal ($J{=}0.111$) and
Emotion-DistilRoBERTa-normal ($J{=}0.111$), both involving the normal
class, consistent with its higher base-rate prevalence providing more stable
attribution signal.  For Kaggle-to-Reddit comparisons,
$8$ of $16$ pairs show $J{=}0.000$, with non-zero values ranging from
$0.053$ to $0.176$.

\begin{table*}[t]
  \centering
  \caption{Cross-platform Jaccard similarity ($J$) of top-$K{=}10$
    gradient-saliency attribution token sets for each of $16$ model-class
    pairs ($4$ models $\times$ $4$ classes).  $J{=}0$ denotes complete token
    vocabulary divergence at $K{=}10$; $J{=}1$ denotes complete identity.
    Random baseline: $J{\approx}0.0001$ for a $50{,}265$-token vocabulary.
    $^\dagger$Non-independent Reddit evaluation (Section~\ref{sec:models}).}
  \label{tab:jaccard}
  \renewcommand{\arraystretch}{1.25}
  \begin{tabular}{@{}llccc@{}}
    \toprule
    \textbf{Model} & \textbf{Class}
      & \textbf{$J$ (Kaggle$\to$Reddit)}
      & \textbf{$J$ (Kaggle$\to$Twitter)}
      & \textbf{$J$ (Reddit$\to$Twitter)} \\
    \midrule
    BERT & Normal     & \textbf{0.000} & \textbf{0.000} & 0.053 \\
    BERT & Depression & 0.053          & \textbf{0.000} & 0.053 \\
    BERT & Anxiety    & \textbf{0.000} & \textbf{0.000} & 0.053 \\
    BERT & Stress     & \textbf{0.000} & \textbf{0.000} & 0.111 \\
    \midrule
    RoBERTa & Normal     & 0.053          & 0.111          & 0.111 \\
    RoBERTa & Depression & \textbf{0.000} & \textbf{0.000} & 0.053 \\
    RoBERTa & Anxiety    & 0.176          & \textbf{0.000} & 0.111 \\
    RoBERTa & Stress     & \textbf{0.000} & \textbf{0.000} & 0.111 \\
    \midrule
    Emotion-DistilRoBERTa & Normal     & 0.053          & 0.111          & 0.053 \\
    Emotion-DistilRoBERTa & Depression & 0.053          & \textbf{0.000} & \textbf{0.000} \\
    Emotion-DistilRoBERTa & Anxiety    & \textbf{0.000} & \textbf{0.000} & \textbf{0.000} \\
    Emotion-DistilRoBERTa & Stress     & 0.053          & \textbf{0.000} & 0.053 \\
    \midrule
    GoEmotions-RoBERTa$^\dagger$ & Normal     & 0.053          & \textbf{0.000} & 0.111 \\
    GoEmotions-RoBERTa$^\dagger$ & Depression & \textbf{0.000} & \textbf{0.000} & 0.111 \\
    GoEmotions-RoBERTa$^\dagger$ & Anxiety    & \textbf{0.000} & \textbf{0.000} & 0.111 \\
    GoEmotions-RoBERTa$^\dagger$ & Stress     & 0.053          & \textbf{0.000} & 0.053 \\
    \bottomrule
  \end{tabular}
\end{table*}

\begin{figure}[t]
  \centering
  \includegraphics[width=\columnwidth]{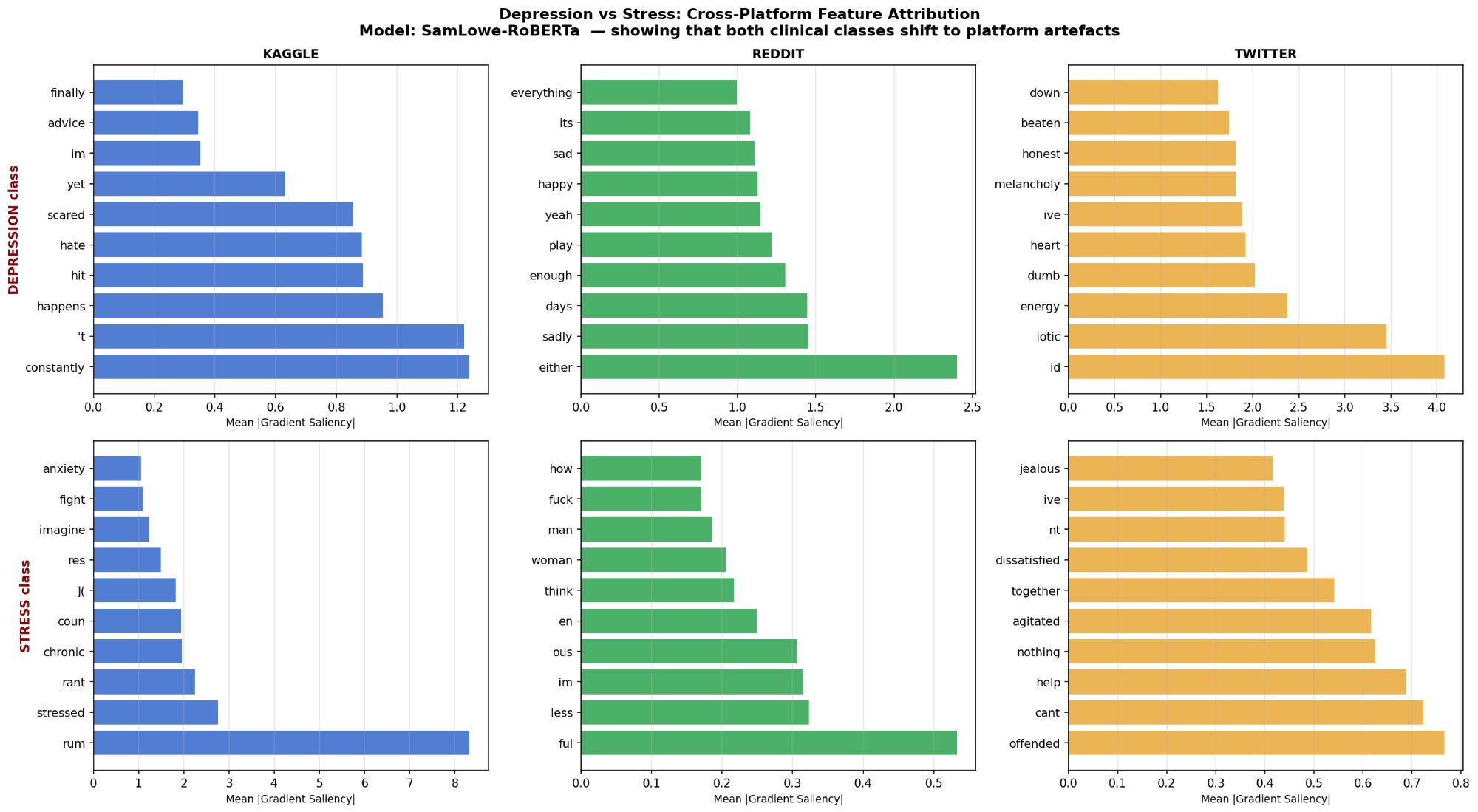}
  \caption{Top-15 gradient saliency token attributions for
    GoEmotions-RoBERTa, Depression class (top) and Stress class (bottom),
    across Kaggle (blue), Reddit (green), and Twitter (orange).  Token
    scores are normalised per platform.}
  \label{fig:attr_goemotions}
\end{figure}

\begin{figure}[t]
  \centering
  \includegraphics[width=\columnwidth]{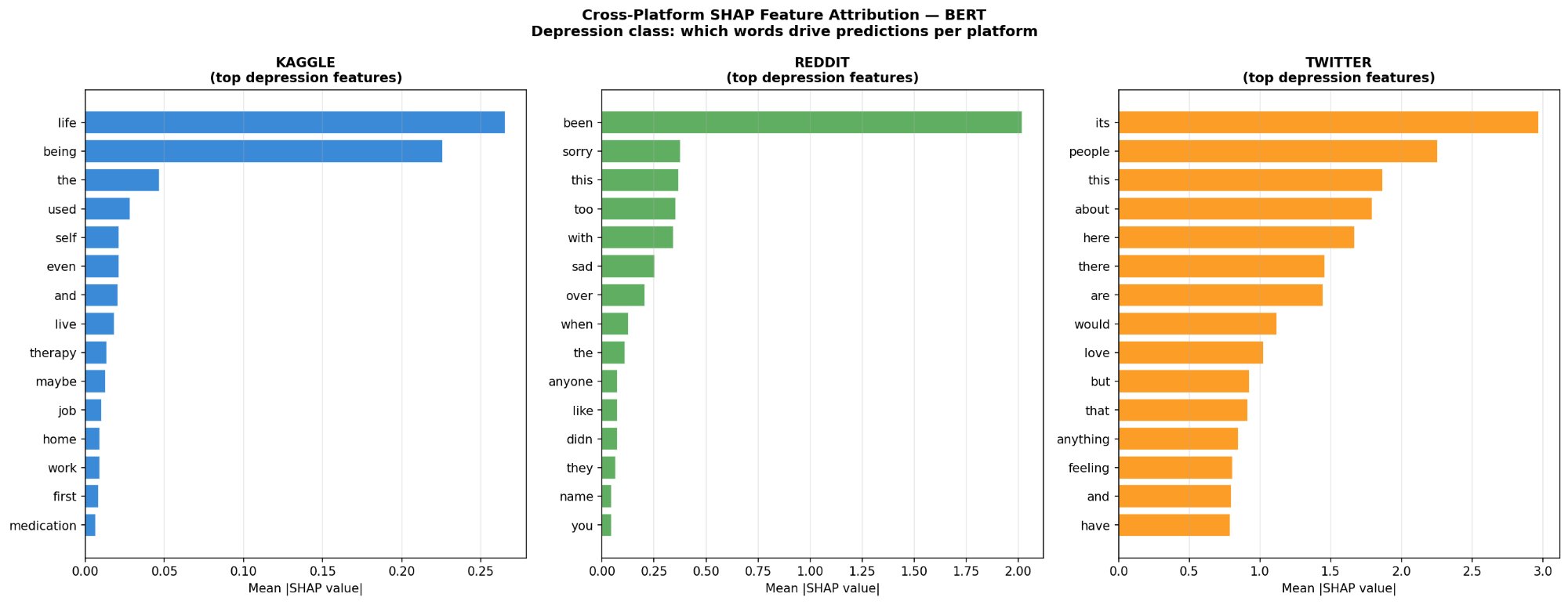}
  \caption{Top-15 gradient saliency token attributions for BERT,
    Depression class, across Kaggle, Reddit, and Twitter.  Token scores
    are normalised per platform.}
  \label{fig:attr_bert}
\end{figure}

\begin{figure}[t]
  \centering
  \includegraphics[width=\columnwidth]{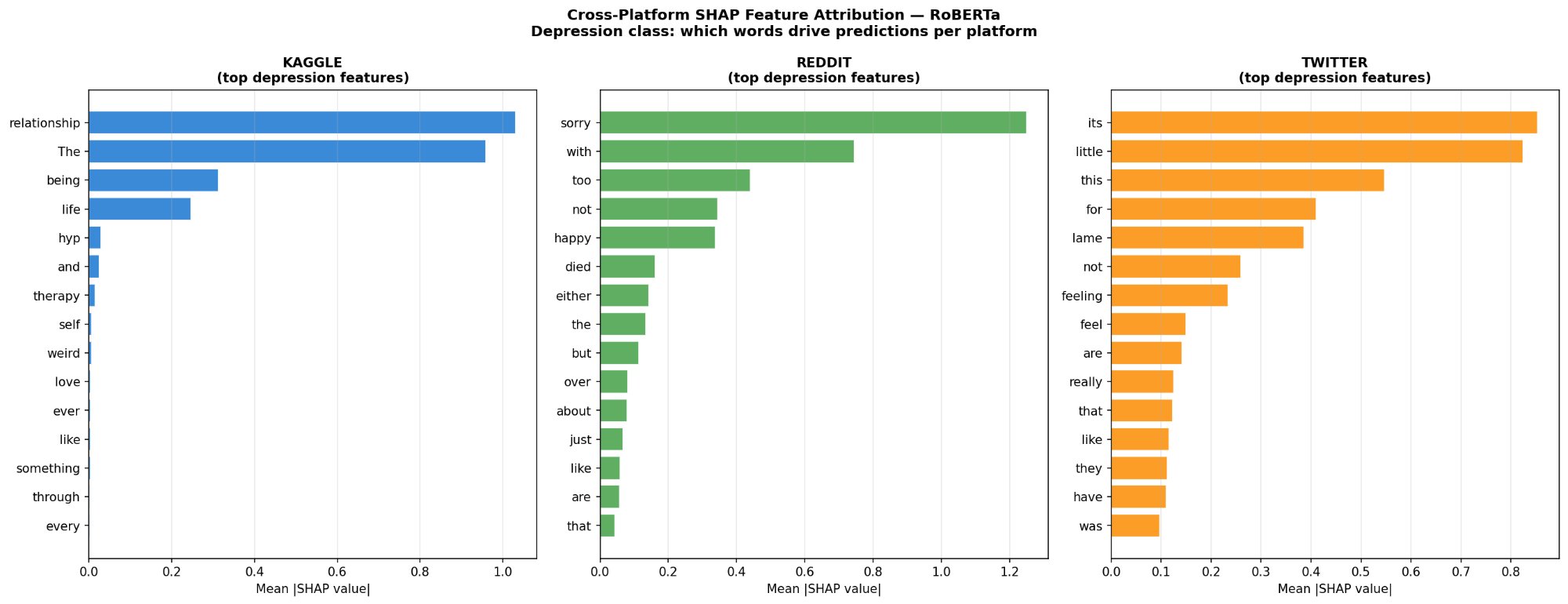}
  \caption{Top-15 gradient saliency token attributions for RoBERTa,
    Depression class, across Kaggle, Reddit, and Twitter.  Token scores
    are normalised per platform.}
  \label{fig:attr_roberta}
\end{figure}

Proxy-clinical signal retention analysis --- using a reference vocabulary
of $45$ terms drawn from DSM-5 symptom descriptions and established
clinical lexicons --- shows that top-10 model features match this vocabulary
at rates of $0$--$30\%$ on Kaggle but uniformly at $0\%$ on Reddit and
Twitter.  On Kaggle, model attention concentrates on proxy-clinical
vocabulary consistent with DSM-adjacent terminology.  On Reddit and Twitter,
attribution patterns shift predominantly toward platform-specific
conversational vocabulary
(see Figs.~\ref{fig:attr_goemotions}--\ref{fig:attr_roberta}).
(Gradient saliency has known limitations; near-zero Jaccard values should
be interpreted as consistent with attribution instability rather than as
definitive attribution collapse --- see Section~\ref{sec:limitations},
item~3.)

\subsection{Fine-Tuning vs.\ Temperature Scaling}
\label{sec:ft_vs_ts}

\begin{table*}[t]
  \centering
  \caption{Target-domain fine-tuning vs.\ temperature scaling comparison.
    Each model was fine-tuned for three additional epochs on the same $10\%$
    stratified calibration split.  \auc{} and \ece{} reported on the
    remaining $90\%$ of the target-platform test set.
    $\Delta$AUC $=$ fine-tuned AUC $-$ baseline cross-platform AUC.
    Mean $\Delta$AUC across all model--platform pairs: $+0.216$.
    Fine-tuning conducted under single seed (seed = 42).
    Base AUC values are single-seed (seed\,${=}$\,42) results for
    comparability with fine-tuned AUC; five-seed mean AUC values for the
    same models appear in Table~\ref{tab:cross_platform}.
    $^\dagger$Non-independent Reddit evaluation (Section~\ref{sec:models}).
    $^\ddagger$Emotion-DistilRoBERTa Twitter fine-tuned AUC of $0.961$
    should be interpreted with caution given the small sample
    ($n_{\rm cal}{=}286$) and single-seed design.}
  \label{tab:finetuning}
  \renewcommand{\arraystretch}{1.25}
  \begin{tabular}{@{}llccccccc@{}}
    \toprule
    \textbf{Model} & \textbf{Platform}
      & $n_{\rm cal}$ & $n_{\rm eval}$
      & \textbf{Base AUC} & \textbf{FT AUC}
      & $\bm{\Delta}$\textbf{AUC}
      & \textbf{FT ECE} & \textbf{TS ECE} \\
    \midrule
    BERT                      & Reddit  & 625 & 5,632 & 0.642 & 0.819 & $+0.177$ & 0.099 & 0.047 \\
    BERT                      & Twitter & 286 & 2,597 & 0.598 & 0.770 & $+0.172$ & 0.074 & 0.033 \\
    RoBERTa                   & Reddit  & 625 & 5,632 & 0.634 & 0.862 & $+0.228$ & 0.082 & 0.037 \\
    RoBERTa                   & Twitter & 286 & 2,597 & 0.606 & 0.805 & $+0.198$ & 0.061 & 0.007 \\
    Emotion-DistilRoBERTa     & Reddit  & 625 & 5,632 & 0.681 & 0.846 & $+0.166$ & 0.082 & 0.033 \\
    Emotion-DistilRoBERTa$^\ddagger$ & Twitter & 286 & 2,597 & 0.611 & 0.961 & $+0.350$ & 0.019 & 0.043 \\
    GoEmotions-RoBERTa$^\dagger$  & Reddit  & 625 & 5,632 & 0.701 & 0.925 & $+0.224$ & 0.046 & 0.038 \\
    GoEmotions-RoBERTa$^\dagger$  & Twitter & 286 & 2,597 & 0.608 & 0.822 & $+0.215$ & 0.094 & 0.037 \\
    \midrule
    \textbf{Mean}             &         &     &       &       &       & $\bm{+0.216}$ & & \\
    \bottomrule
  \end{tabular}
\end{table*}

\begin{figure}[t]
  \centering
  \includegraphics[width=\columnwidth]{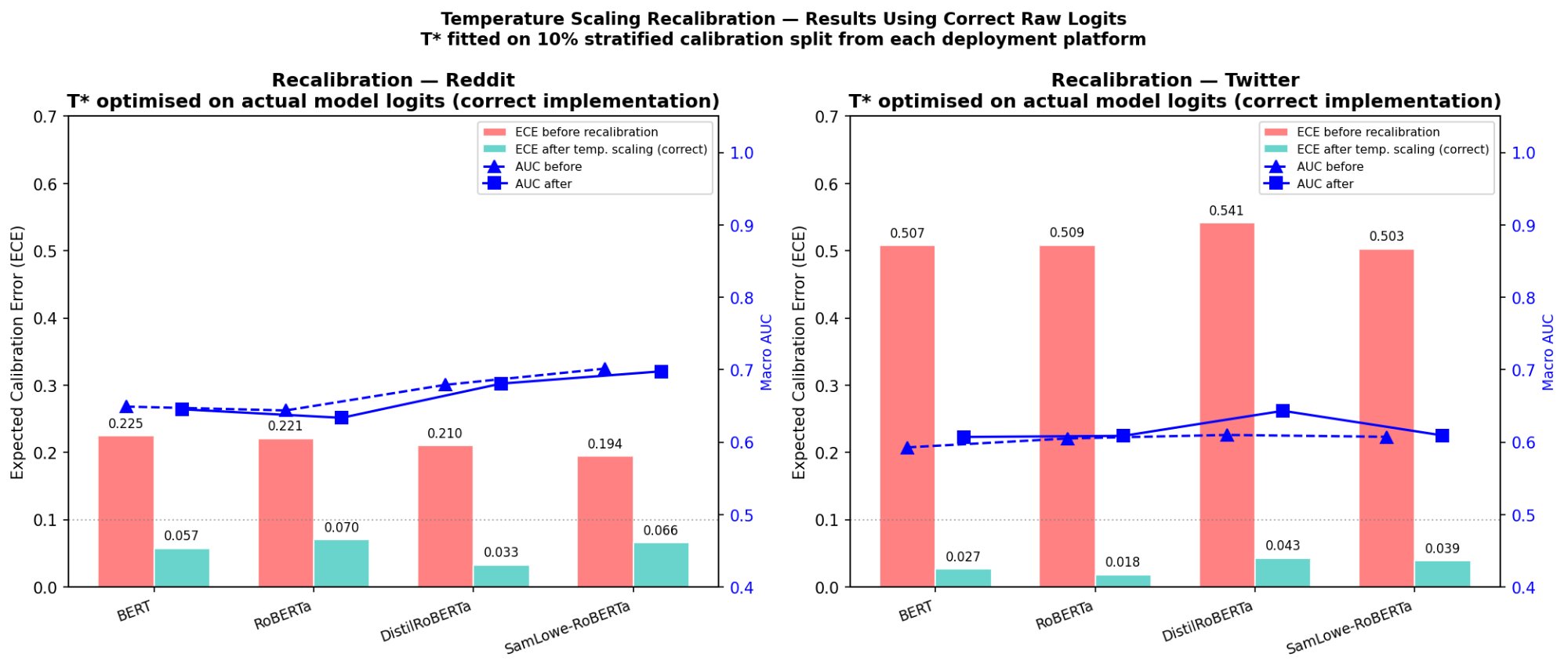}
  \caption{Temperature scaling vs.\ target-domain fine-tuning as
    calibration remediation strategies for all four models on Reddit and
    Twitter.}
  \label{fig:ft_vs_ts}
\end{figure}

In this single-seed fine-tuning experiment, mean \auc{} improved by $0.216$
across model--platform pairs (Table~\ref{tab:finetuning};
Fig.~\ref{fig:ft_vs_ts}).  The Emotion-DistilRoBERTa Twitter result
($\Delta$\auc{} $= {+}0.350$; $n_{\rm cal}{=}286$) inflates this mean;
excluding that pair gives a mean gain of $0.197$.  These results should be
treated as indicative rather than definitive given the single-seed design.
Where labelled target-platform samples are available, fine-tuning recovers
discriminative performance substantially more effectively than post-hoc
recalibration.
Temperature scaling remains appropriate when fine-tuning is computationally
infeasible or when only calibration correction --- rather than
discriminative recovery --- is the operational goal.

\subsection{Sensitivity Analysis}
\label{sec:sens_results}

\begin{figure}[t]
  \centering
  \includegraphics[width=\columnwidth]{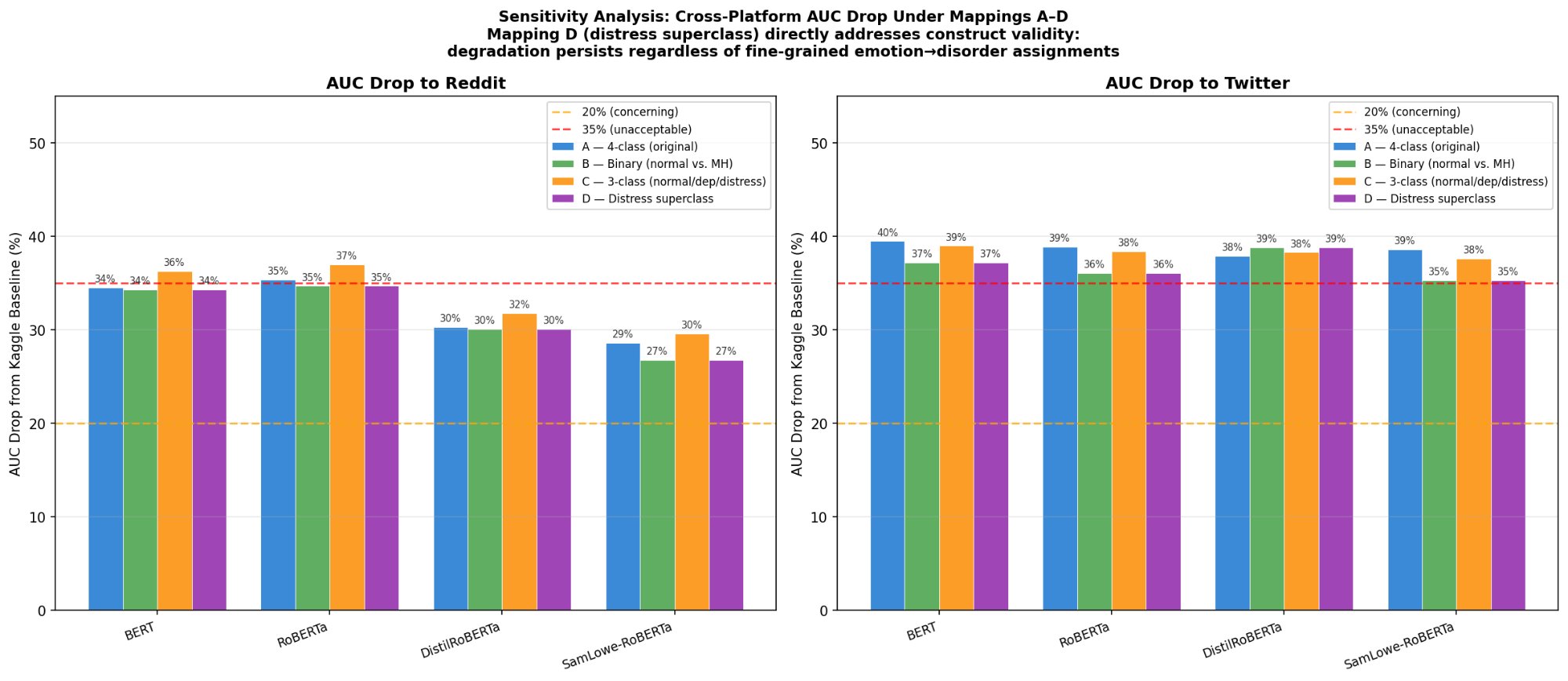}
  \caption{Cross-platform \auc{} under four label mapping schemes.  Each
    bar shows mean \auc{} with $95\%$ bootstrap confidence interval.}
  \label{fig:sensitivity}
\end{figure}

Cross-platform degradation patterns are consistent under all four
alternative label mapping schemes (Table~\ref{tab:sensitivity};
Fig.~\ref{fig:sensitivity}).  Under Mapping~B (binary), which eliminates
all fine-grained class-specific label ambiguity, Reddit drops remain
$26.8$--$34.7\%$ and Twitter drops remain $35.3$--$39.0\%$.  The
consistency of cross-platform \auc{} degradation across all four
alternative mapping schemes --- including Mapping~D, which collapses
anxiety and stress into a single distress superclass --- provides the
primary empirical basis for concluding that generalisation failure is not
an artefact of the label-mapping heuristic.

\begin{table*}[t]
  \centering
  \caption{Cross-platform \auc{} degradation under four label mapping
    schemes: (A) 4-class primary; (B) binary (normal vs.\ any mental health
    condition); (C) 3-class (normal/depression/distress); (D) distress
    superclass (collapsing anxiety and stress).}
  \label{tab:sensitivity}
  \renewcommand{\arraystretch}{1.2}
  \begin{tabular}{@{}llrr@{}}
    \toprule
    \textbf{Model} & \textbf{Mapping}
      & \textbf{Reddit $\Delta$AUC\%}
      & \textbf{Twitter $\Delta$AUC\%} \\
    \midrule
    BERT & A --- 4-class (primary)            & $-34.5\%$ & $-39.5\%$ \\
    BERT & B --- Binary (normal vs.\ MH)      & $-34.3\%$ & $-37.2\%$ \\
    BERT & C --- 3-class (normal/dep/distress)& $-36.3\%$ & $-39.0\%$ \\
    BERT & D --- Distress superclass          & $-34.3\%$ & $-37.2\%$ \\
    \midrule
    RoBERTa & A --- 4-class (primary)           & $-35.4\%$ & $-38.9\%$ \\
    RoBERTa & B --- Binary (normal vs.\ MH)     & $-34.7\%$ & $-36.1\%$ \\
    RoBERTa & C --- 3-class                     & $-37.0\%$ & $-38.4\%$ \\
    RoBERTa & D --- Distress superclass         & $-34.7\%$ & $-36.1\%$ \\
    \midrule
    Emotion-DistilRoBERTa & A --- 4-class       & $-30.3\%$ & $-37.9\%$ \\
    Emotion-DistilRoBERTa & B --- Binary        & $-30.1\%$ & $-38.8\%$ \\
    Emotion-DistilRoBERTa & C --- 3-class       & $-31.8\%$ & $-38.3\%$ \\
    Emotion-DistilRoBERTa & D --- Distress      & $-30.1\%$ & $-38.8\%$ \\
    \midrule
    GoEmotions-RoBERTa & A --- 4-class          & $-28.6\%$ & $-38.6\%$ \\
    GoEmotions-RoBERTa & B --- Binary           & $-26.8\%$ & $-35.3\%$ \\
    GoEmotions-RoBERTa & C --- 3-class          & $-29.6\%$ & $-37.6\%$ \\
    GoEmotions-RoBERTa & D --- Distress         & $-26.8\%$ & $-35.3\%$ \\
    \bottomrule
  \end{tabular}
\end{table*}

\subsection{Truncation and Text Length}
\label{sec:truncation}

Kaggle samples have a mean word count of $118.3$ (median $67.0$;
$p_{75}{=}154.0$) compared with $13.5$ (Reddit) and $19.0$ (Twitter).
An estimated $59.2\%$ of Kaggle samples exceed the $64$-token maximum
sequence length ($n{=}4{,}508$), versus $0.0\%$ of Reddit and $2.8\%$ of
Twitter samples.  Models trained on truncated long-form Kaggle text have
been exposed to clinical narrative structure --- symptom trajectories,
temporal references, and first-person disclosures --- that is structurally
absent from short-form Reddit comments and tweets.  This platform asymmetry
supports the domain-shift interpretation: the cross-platform failure
reflects genuine distributional divergence in communicative genre and text
register --- a form of covariate shift --- rather than merely a consequence
of label-mapping heuristics.

\section{Discussion}
\label{sec:discussion}

Across all five \cpfe{} axes, the results converge on a structural
interpretation: the observed cross-platform failure patterns are not
specific to any single evaluated transformer architecture, and are
consistent with a general property of single-platform proxy-label training.
These failures co-occur across all three independently evaluated models
--- degraded discrimination, miscalibration, attribution shift, and
prediction rate disparities manifest simultaneously on both cross-platform
test sets --- and this consistent co-occurrence across architecturally
diverse models is the study's central empirical observation.

\subsection{The Scale and Consistency of Cross-Platform Failure}

Moving from \auc{} $0.983$--$0.987$ to \auc{} $0.596$--$0.703$ represents
a severe loss of discriminative utility.  \auc{} values of $0.596$--$0.703$
are above chance ($\text{\auc{}}=0.50$) and retain some signal, but fall
below the $\text{\auc{}} \ge 0.80$ threshold commonly cited as a minimum
for discriminative utility in clinical classification tasks~\cite{metz1978,zweig1993}
--- though that benchmark was established with validated clinical outcomes,
not proxy labels.  The F1-macro collapse from $0.862$--$0.883$ to
$0.284$--$0.332$ means that even a basic recall-precision balance across
classes is unachievable on cross-platform data.  \ece{} values of
$0.499$--$0.542$ on Twitter indicate that model-expressed probabilities
deviate substantially from true class probabilities (by approximately $50$
percentage points on average).

The consistency of these findings across three independently evaluated
architecturally diverse models (with GoEmotions-RoBERTa providing a
non-independent partial reference) suggests the failure is not specific to
any particular model, but of the distributional mismatch between
single-platform proxy-label training data and heterogeneous deployment
environments.
As noted, this benchmark was developed for validated clinical endpoints; it
is reported here only to convey the magnitude of degradation relative to a
conventional reference, not as a directly applicable threshold for
proxy-label classification.

Among the four evaluated models, GoEmotions-RoBERTa shows the smallest
Reddit \auc{} drop ($28.6\%$), but this reflects pretraining overlap with
the specific test domain rather than acquired cross-platform robustness;
the advantage does not extend to Twitter (drop $38.6\%$), where all four
models perform similarly.

\subsection{Temperature Scaling as a Practical Post-Hoc Calibration Remedy}
\label{sec:ts_discussion}

Temperature scaling provides an immediate post-hoc response to calibration
failure that requires no model retraining but does require a small labelled
calibration sample from the target platform ($n{=}625$ for Reddit;
$n{=}286$ for Twitter).
As the fine-tuning experiment in Section~\ref{sec:ft_vs_ts} shows, when
labelled target-platform samples are available, model retraining recovers
discriminative performance substantially more effectively than post-hoc
recalibration.  Temperature scaling remains the appropriate remedy when
retraining is computationally infeasible.

\subsection{Construct Validity: Emotion Labels and Clinical Categories}

The cross-platform test sets use emotion-annotated data remapped to clinical
categories.  The construct validity concern --- that anger $\ne$ stress and
fear $\ne$ anxiety clinically --- is directly addressed by Mapping~D
(distress superclass).  Under this mapping, which eliminates the
anxiety/stress distinction and tests only normal vs.\ depression vs.\
distress, degradation is statistically indistinguishable from the primary
analysis on both Reddit (drops $26.8$--$34.3\%$) and Twitter (drops
$35.3$--$39.0\%$).  The observed failure is therefore a property of
cross-platform distribution shift, not of the label mapping.

\subsection{Cross-Platform Deployment Equity Implications}

A model with depression proxy-class $\di{=}0.133$ (RoBERTa on Reddit)
generates depression proxy-class predictions at approximately one-seventh
the rate observed on the Kaggle reference platform.  The
prior-shift-adjusted \di{} of $0.11$--$0.29$ confirms that a substantial
disparity persists beyond what prevalence differences alone explain.  The
\eod{} values of $0.753$--$0.830$ for depression, anxiety, and stress on
Reddit indicate that the conditional true positive rate drops by
$75$--$83$ percentage points relative to within-platform performance.

These findings motivate treating platform-stratified prediction equity metrics
as an informative diagnostic dimension during pre-deployment evaluation, not
as definitive evidence of algorithmic bias.  The appropriate response to
large \di{} and \eod{} values is platform-specific model validation and
potential recalibration.

\subsection{Illustrative CPFE Degradation Benchmarks}

In this study, all three independently evaluated models showed \auc{} drops
exceeding $30\%$ on both platforms, a magnitude associated with near-chance
minority-class performance.  These figures derive from one training corpus,
three platforms, four models, and proxy emotion labels; they should be
treated as descriptive baselines for future cross-platform evaluation work
rather than as validated decision thresholds.

\section{Limitations}
\label{sec:limitations}

\begin{enumerate}[leftmargin=*, label=\textbf{\arabic*.}]

  \item \textbf{Construct validity of cross-platform labels.}
  The emotion$\to$clinical label mappings are heuristic approximations.
  While Mapping~D provides evidence of robustness to the most salient
  construct validity concern, human-annotated clinical labels on Reddit and
  Twitter would be required to fully disentangle construct mismatch from
  distributional failure.

  \item \textbf{Pretraining contamination.}
  GoEmotions-RoBERTa and Emotion-DistilRoBERTa have pretraining overlap with
  the evaluation domains.  Their Reddit performance figures are partially
  non-independent and should be treated as upper bounds.

  \item \textbf{Attribution method reliability.}
  Token importance was estimated by gradient-based saliency~\cite{simonyan2014},
  a local linear approximation to feature importance.  This method has known
  limitations: it can be sensitive to input normalisation and random seed,
  and its faithfulness to model decision logic is not
  guaranteed~\cite{adebayo2018,jain2019}.  Multiple studies have demonstrated
  that gradient saliency rankings can be inconsistent across seeds and are
  not always faithful to model decision logic.  The near-zero Jaccard values
  should be interpreted as consistent with substantial attribution
  instability rather than as definitive evidence of attribution collapse.
  Confirmation with Shapley-based or Integrated Gradients
  methods~\cite{sundararajan2017} is a priority for future work.

  \item \textbf{Class imbalance and prior shift.}
  The Kaggle partition is heavily skewed toward depression ($56.6\%$)
  relative to Reddit ($6.6\%$) and Twitter ($30.1\%$).  Separate estimation
  of prior shift vs.\ likelihood shift contributions would strengthen causal
  attribution of the observed cross-platform prediction equity disparities.

  \item \textbf{Residual variance in secondary metrics across seeds.}
  While \auc{} degradation was consistent across all five seeds, secondary
  metrics --- specifically Symmetric Disparate Impact, Equalized Odds
  Difference, and Jaccard similarity scores --- were computed only for the
  primary seed (seed\,${=}$\,42); seed-level variance for these metrics was
  not characterised.  Multi-seed replication of equity and attribution
  metrics is a priority for future work.

  \item \textbf{Fine-tuning comparison.}
  The mean \auc{} gain of $0.216$ is a point estimate without characterised
  variance (single seed; seed~$= 42$).  Multi-seed replication of the
  fine-tuning experiment is a priority for future work.

  \item \textbf{Temperature scaling calibration split from test data.}
  The optimal temperature $T^*$ was estimated on a $10\%$ stratified split
  from the target platform test set; post-calibration \ece{} values are
  therefore potentially optimistic relative to a fully novel deployment
  sample.

  \item \textbf{Absence of non-transformer and zero-shot baselines.}
  It is unknown whether classical models (logistic regression, SVM) or large
  zero-shot language models (GPT-series) exhibit the same cross-platform
  degradation patterns.  These comparisons would help establish whether the
  observed failure is specific to fine-tuned transformers or a more general
  property of cross-platform text classification.

\end{enumerate}

\section{Conclusion}
\label{sec:conclusion}

The \cpfe{} framework --- a five-axis audit covering discriminative
performance, calibration, statistical significance, prediction equity, and
attribution stability --- was applied to four transformer-based classifiers
trained on a Kaggle mental health corpus and evaluated cross-platform on
Reddit and Twitter using remapped proxy labels (see Section~\ref{sec:data}).
Across three independently evaluated models (BERT, RoBERTa,
Emotion-DistilRoBERTa) and five training seeds, all five axes revealed
co-occurring failures:
\auc{} fell $30.3$--$35.4\%$ on Reddit and $37.9$--$39.5\%$ on Twitter;
\ece{} rose from $0.056$--$0.060$ within-platform to $0.499$--$0.542$ on
Twitter; gradient attribution token overlap was near-zero ($J{=}0$ in
$14/16$ model--class pairs at $K{=}10$); and cross-platform prediction rate
disparities severely violated the four-fifths rule~\cite{barocas2019}.

Temperature scaling substantially improved calibration (mean \ece{}
reduction $88.0\%$) without restoring discrimination; a single-seed
target-domain fine-tuning experiment on the same $10\%$ labelled split
produced a mean \auc{} gain of $0.216$ (excluding an outlier pair,
$0.197$), suggesting that available target-platform labels may provide
greater benefit as training signal than as calibration signal.
These results were consistent across four label mapping schemes and five
independent training seeds, providing strong empirical support for the view
that cross-platform validation across all five \cpfe{} axes should be
treated as a standard requirement for mental health \textsc{nlp} systems
intended for heterogeneous deployment environments.

\section*{Author Contributions}

Conceptualisation: R.S.P., S.Y.  Methodology: R.S.P., S.Y.  Software:
R.S.P.  Formal analysis: R.S.P., S.Y.  Investigation: R.S.P., S.Y.  Data
curation: R.S.P.  Writing --- original draft: R.S.P.  Writing --- review \&
editing: R.S.P., S.Y.  Visualisation: R.S.P.

\section*{Data and Code Availability}

All analysis code is available at
\url{https://github.com/Rajveer-code/mental-health-fairness-nlp}.
The GoEmotions dataset~\cite{demszky2020} and dair-ai/emotion
dataset~\cite{saravia2018} are publicly available on HuggingFace Datasets.
The Kaggle mental health dataset~\cite{sarkar2022} is available at
\url{https://www.kaggle.com/datasets/suchintikasarkar/sentiment-analysis-for-mental-health}.

\section*{Declaration of Competing Interests}

The authors declare no competing financial or non-financial interests.


\appendices

\section{Supplementary Figures}
\label{app:supplementary}

The following supplementary figures are provided to support the main text.

\begin{figure}[h!]
  \centering
  \includegraphics[width=\columnwidth]{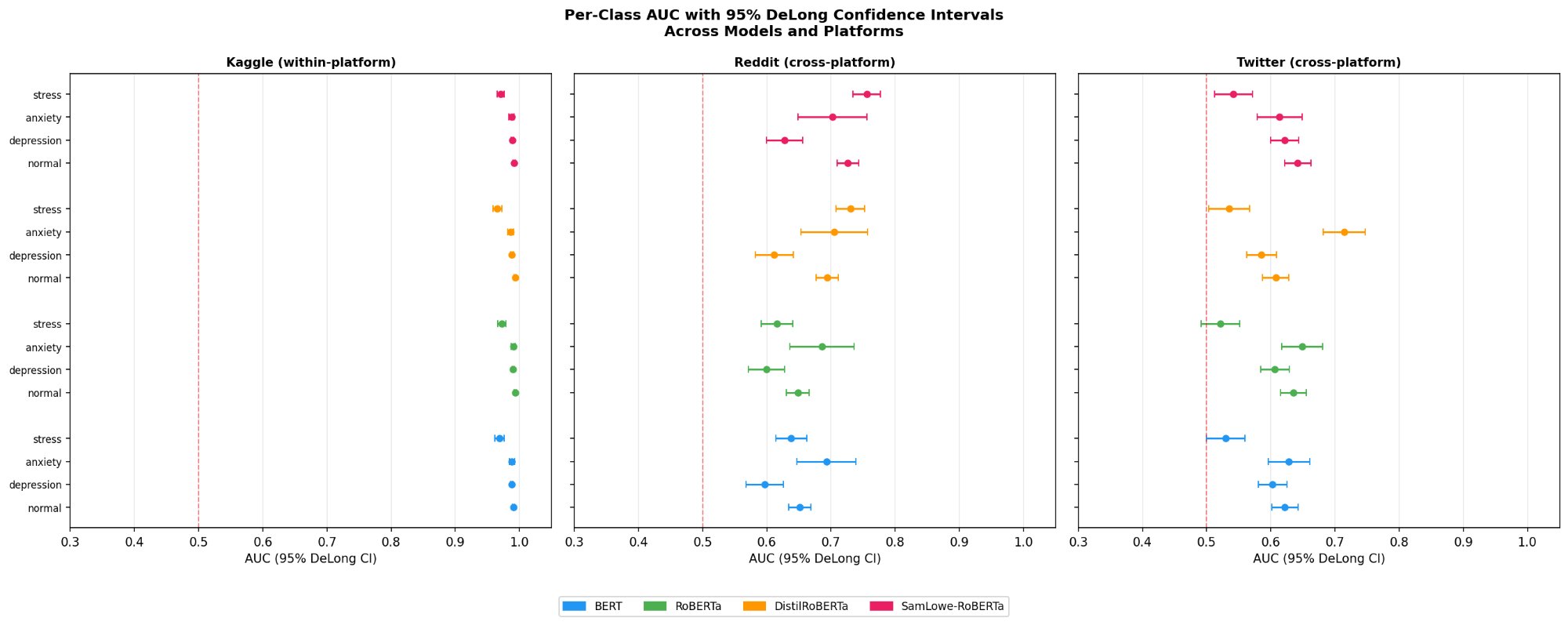}
  \caption{Supplementary Fig.\ S1: Per-class one-vs-rest \auc{} with
    $95\%$ DeLong confidence intervals across all four models and three
    platforms ($48$ estimates).  The red dashed line marks chance
    performance (\auc{} $= 0.50$).  Stress class CIs on Twitter approach
    or touch the chance boundary for BERT ($0.530$ [$0.500$, $0.560$])
    and RoBERTa ($0.522$ [$0.492$, $0.551$]).}
  \label{fig:s1}
\end{figure}

\begin{figure}[h!]
  \centering
  \includegraphics[width=\columnwidth]{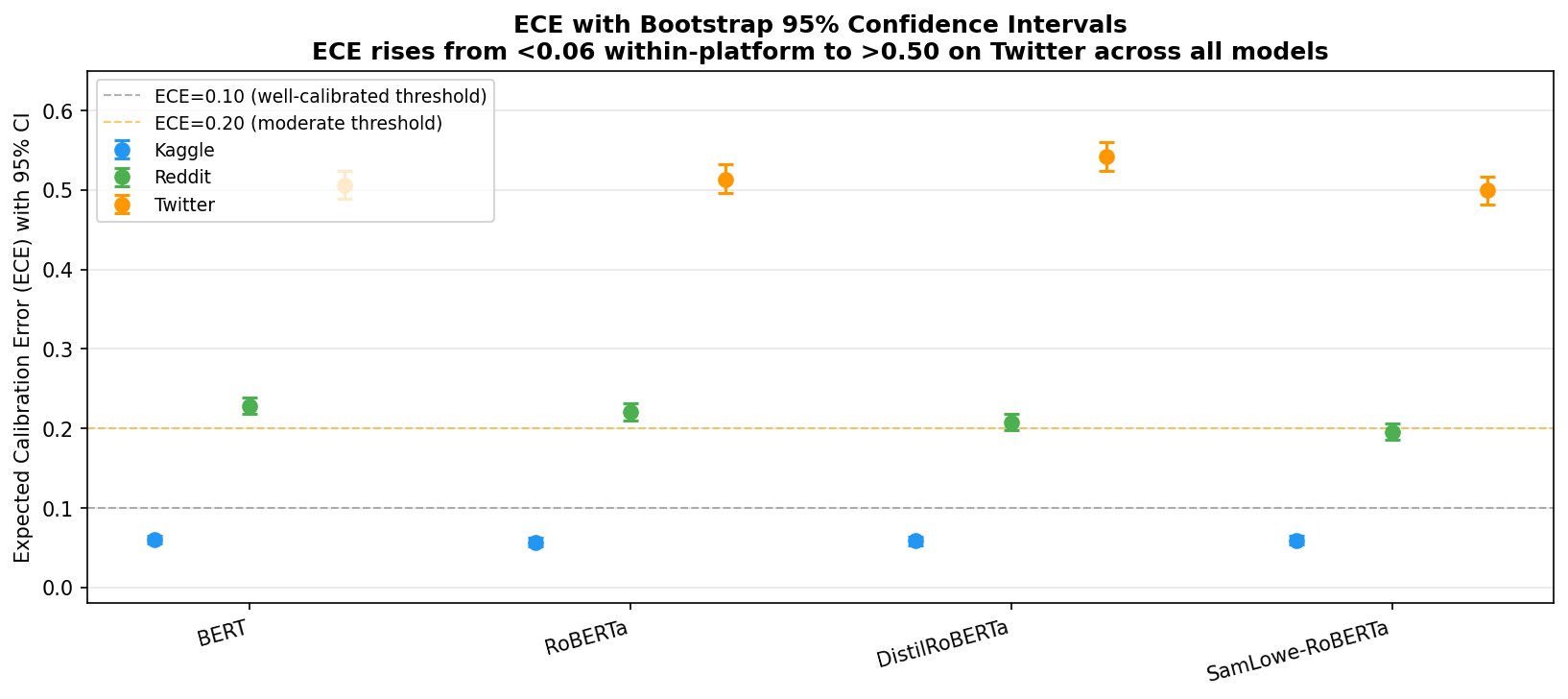}
  \caption{Supplementary Fig.\ S2: Expected Calibration Error (\ece{})
    with $95\%$ bootstrap CIs ($B{=}1{,}000$) across all models and
    platforms.  Within-platform Kaggle \ece{} is consistently below
    $0.06$; Reddit \ece{} is $0.196$--$0.229$; Twitter \ece{} exceeds
    $0.499$--$0.542$ for all models.}
  \label{fig:s2}
\end{figure}

\begin{figure}[h!]
  \centering
  \includegraphics[width=\columnwidth]{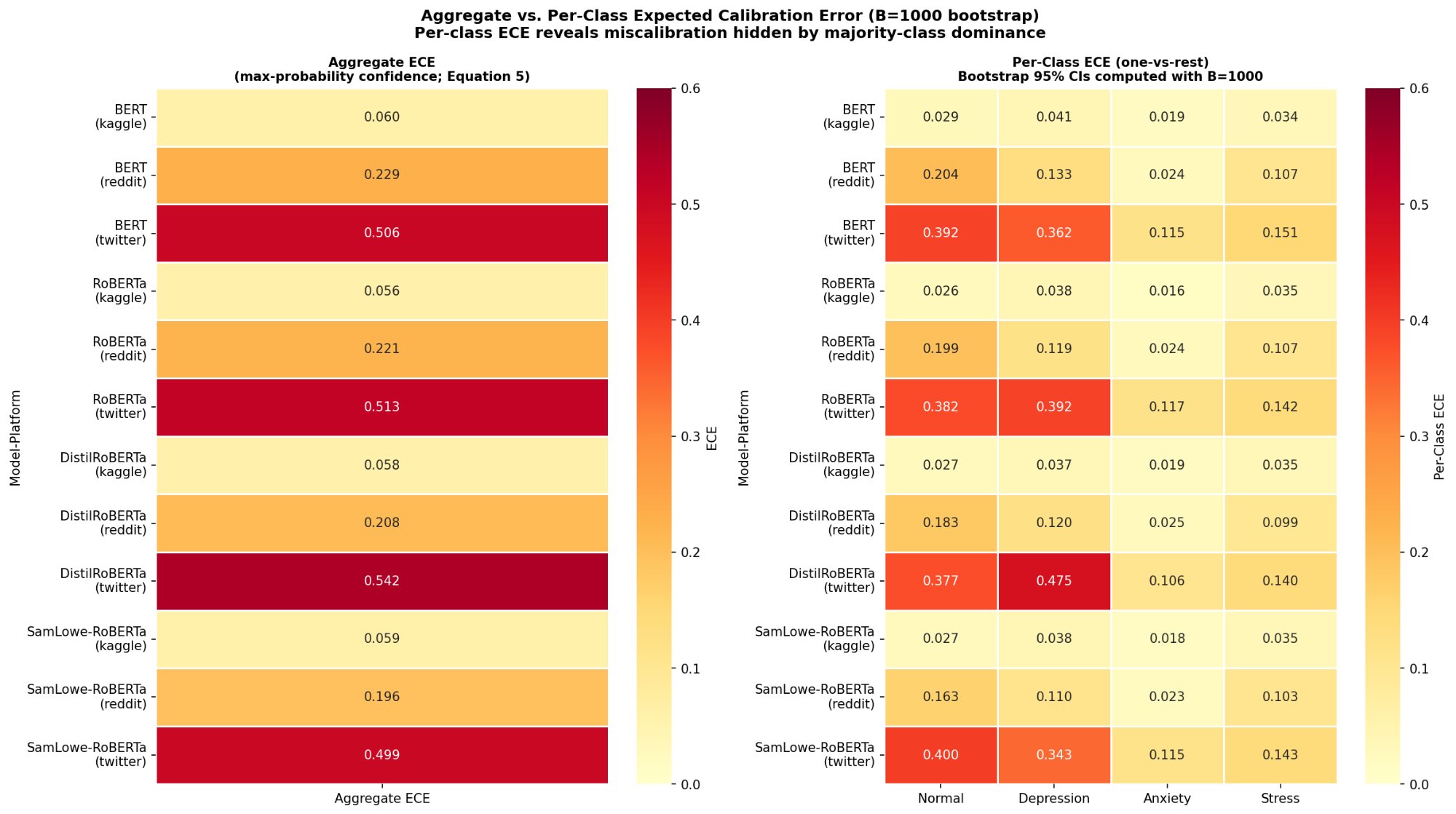}
  \caption{Supplementary Fig.\ S3: Aggregate \ece{} (left panel) and
    per-class one-vs-rest \ece{} with bootstrap $95\%$ CIs (right panel),
    $B{=}1{,}000$.}
  \label{fig:s3}
\end{figure}

\begin{figure}[h!]
  \centering
  \includegraphics[width=\columnwidth]{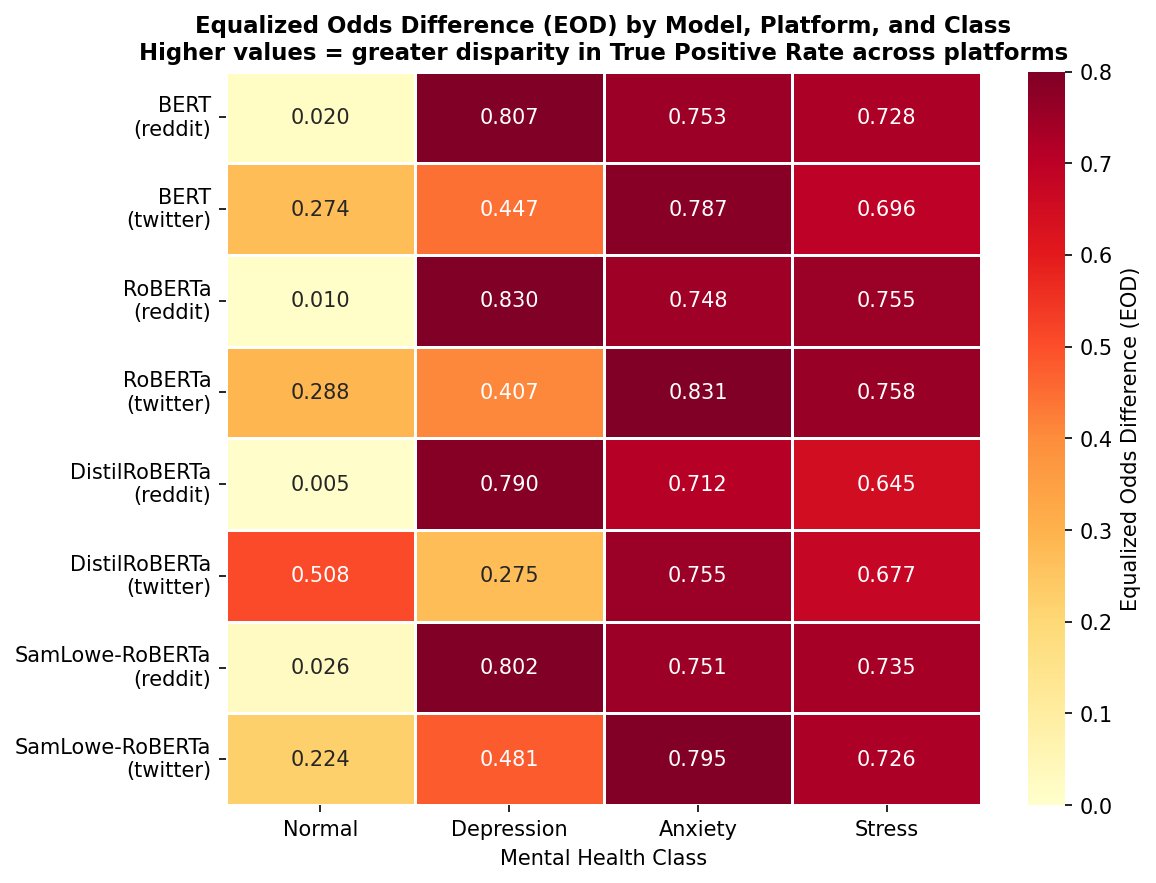}
  \caption{Supplementary Fig.\ S4: Equalized Odds Difference (\eod{})
    heatmap across all four models, platforms, and classes.  \eod{} for
    Depression on Reddit reaches $0.807$--$0.830$ across all models.}
  \label{fig:s4}
\end{figure}

\begin{figure}[h!]
  \centering
  \includegraphics[width=\columnwidth]{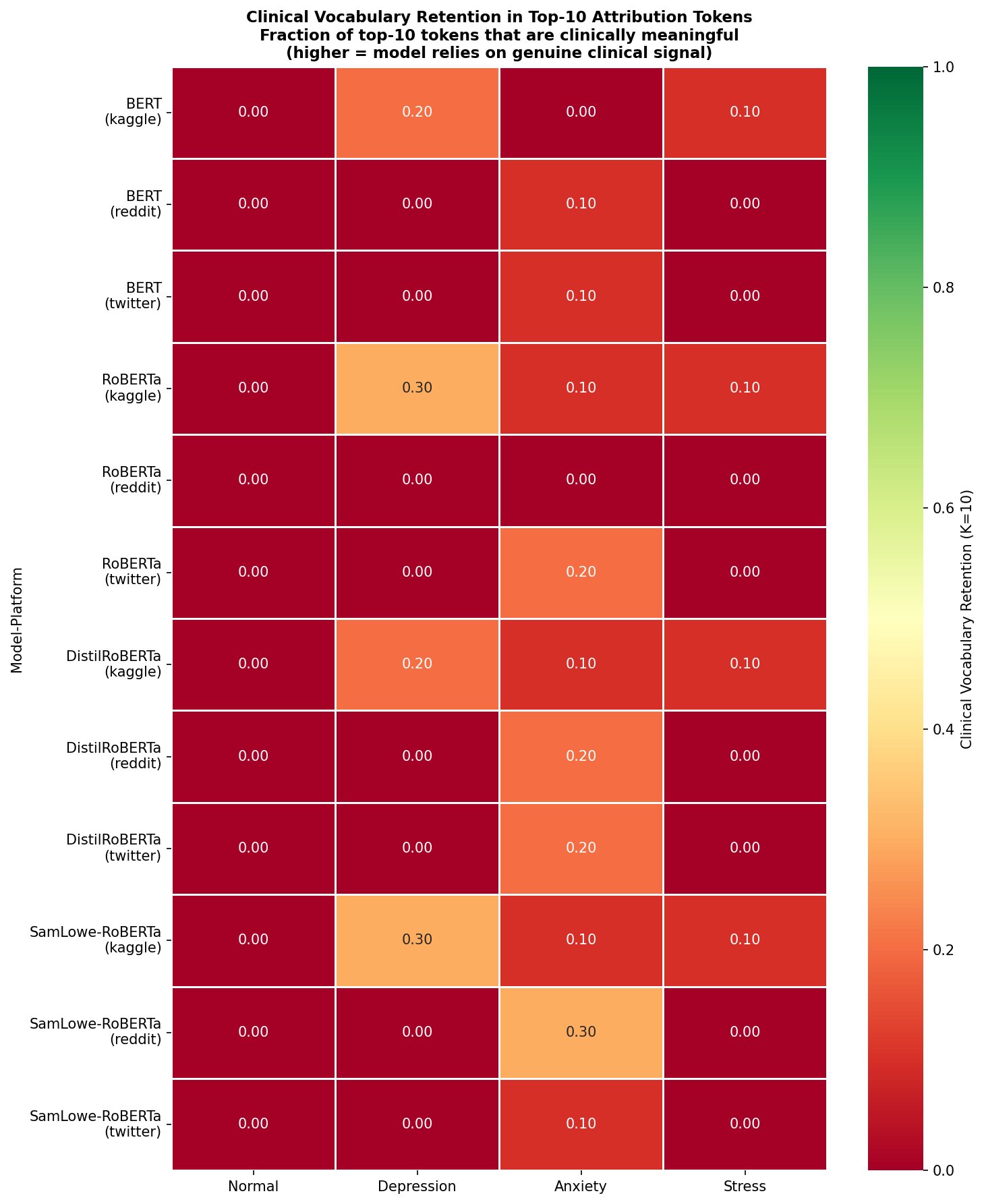}
  \caption{Supplementary Fig.\ S5: Proxy-clinical vocabulary retention
    in top-10 attribution tokens.  Each cell shows the fraction of the
    top-10 gradient saliency tokens that match a predefined expert
    proxy-clinical vocabulary list.}
  \label{fig:s5}
\end{figure}

\begin{figure}[h!]
  \centering
  \includegraphics[width=\columnwidth]{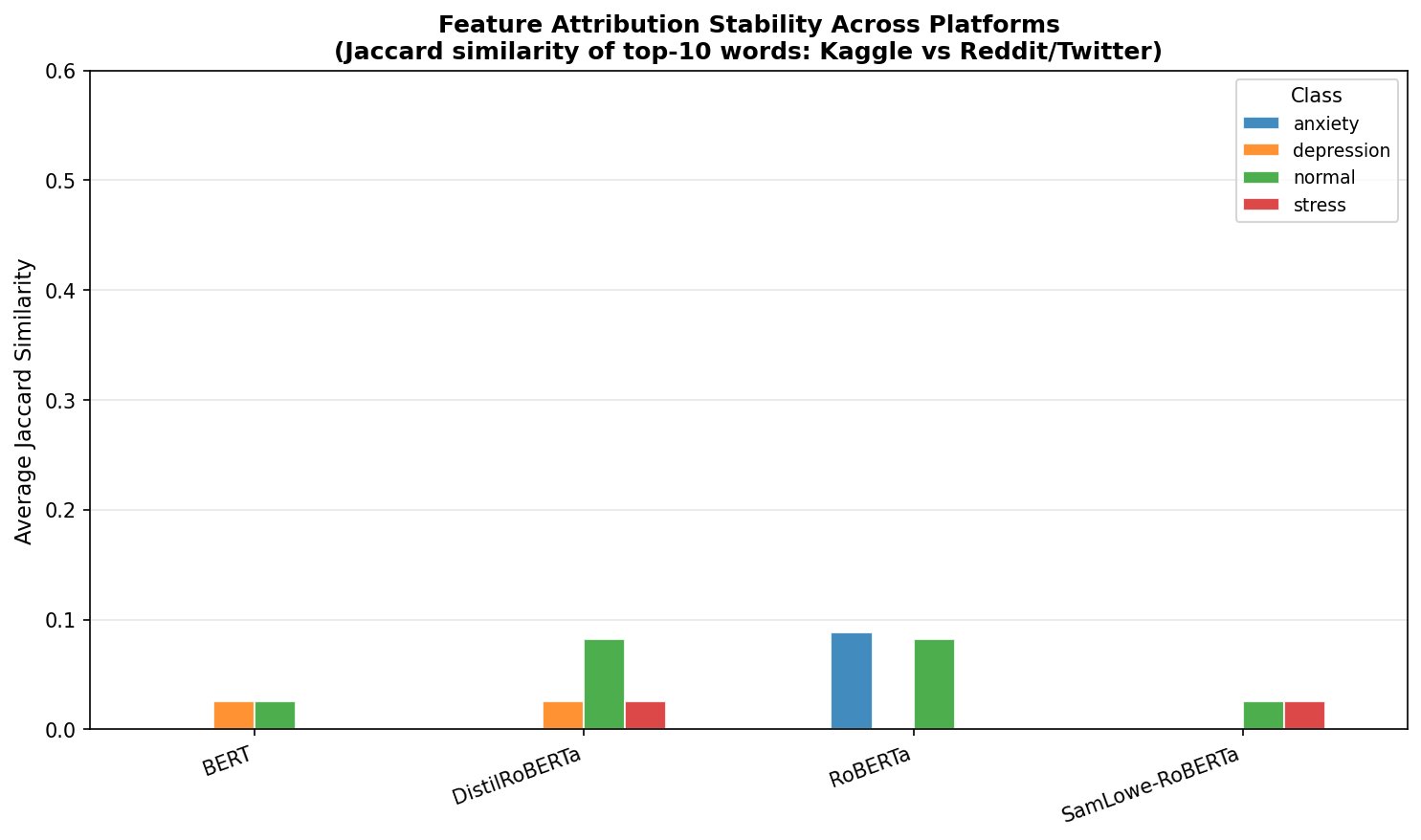}
  \caption{Supplementary Fig.\ S6: Average Jaccard similarity of top-10
    attribution words across both cross-platform comparisons, by model
    and class.}
  \label{fig:s6}
\end{figure}

\begin{figure}[h!]
  \centering
  \includegraphics[width=\columnwidth]{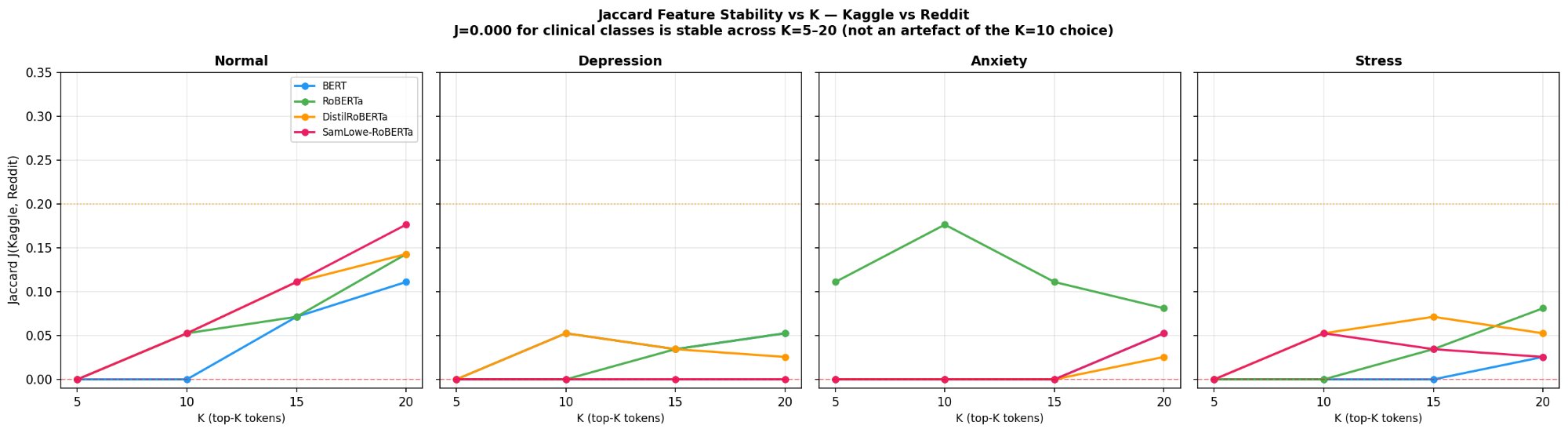}
  \caption{Supplementary Fig.\ S7: Jaccard feature stability sensitivity
    to $K$ (number of top-attributed tokens), for Kaggle$\to$Reddit
    comparison.  Shaded bands show range across four models.}
  \label{fig:s7}
\end{figure}

\begin{figure}[h!]
  \centering
  \includegraphics[width=\columnwidth]{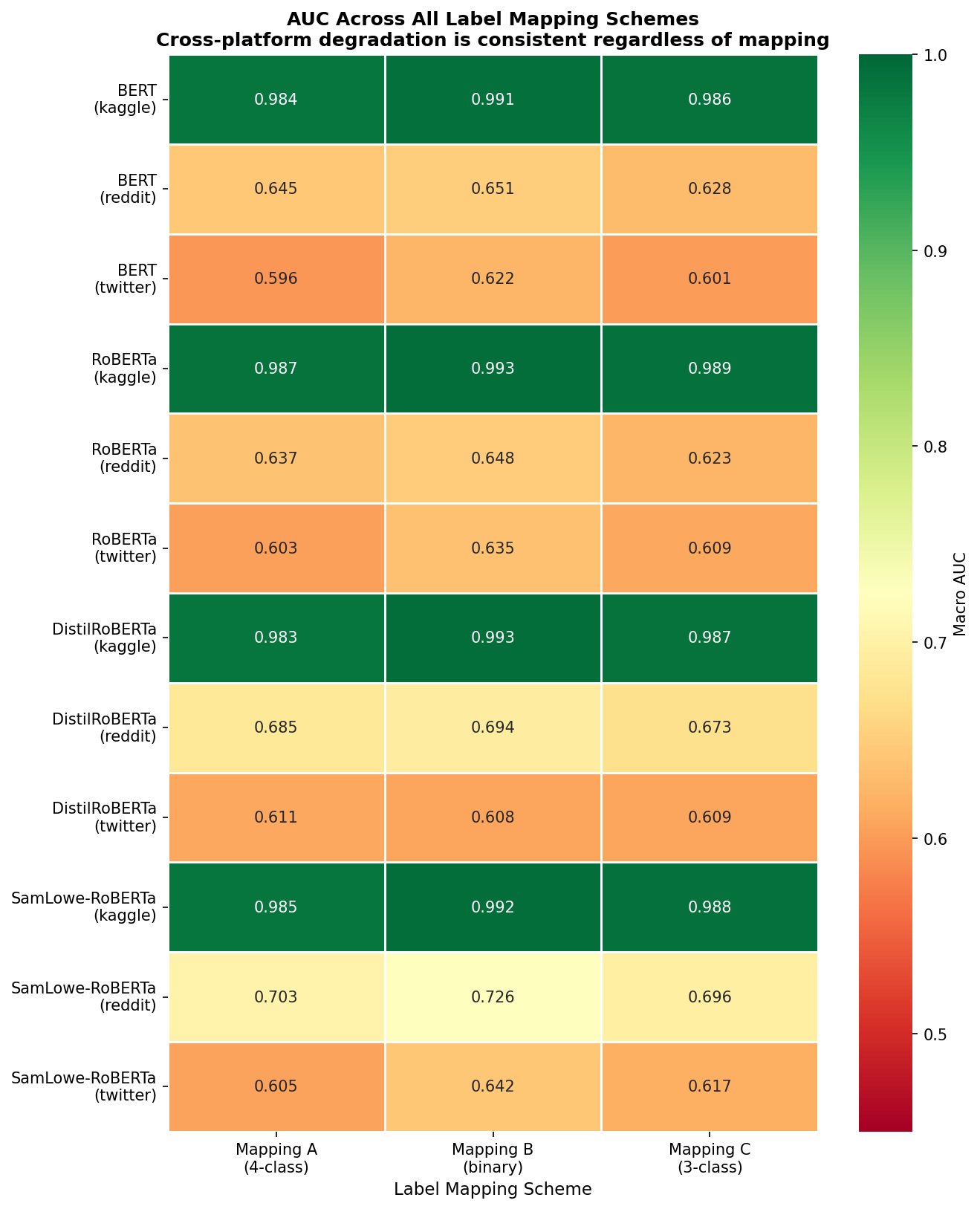}
  \caption{Supplementary Fig.\ S8: Macro \auc{} heatmap across all
    models, platforms, and label mapping schemes (A--D).}
  \label{fig:s8}
\end{figure}

\begin{figure}[h!]
  \centering
  \includegraphics[width=\columnwidth]{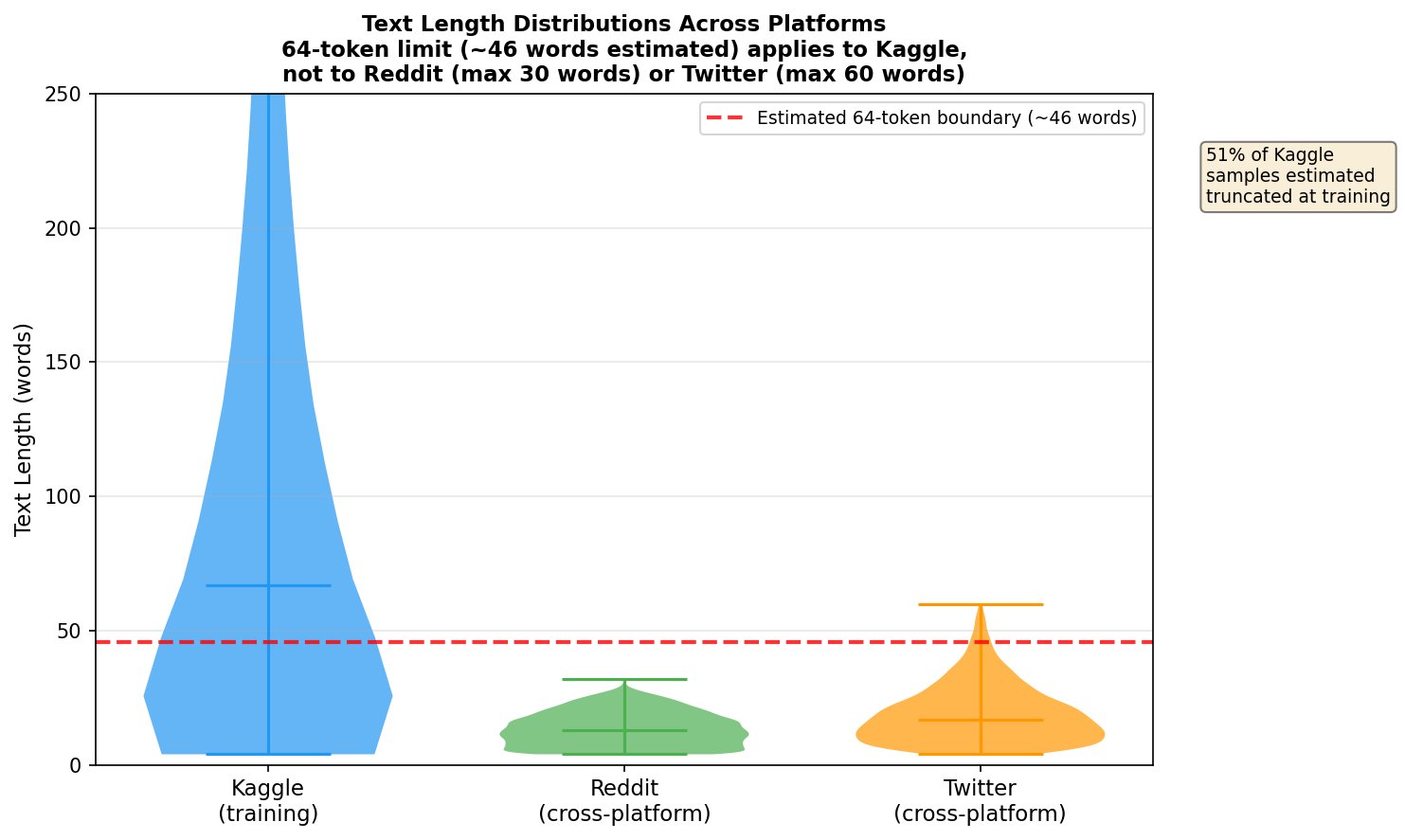}
  \caption{Supplementary Fig.\ S9: Text length distributions across
    platforms (violin plots).}
  \label{fig:s9}
\end{figure}

\begin{figure}[h!]
  \centering
  \includegraphics[width=\columnwidth]{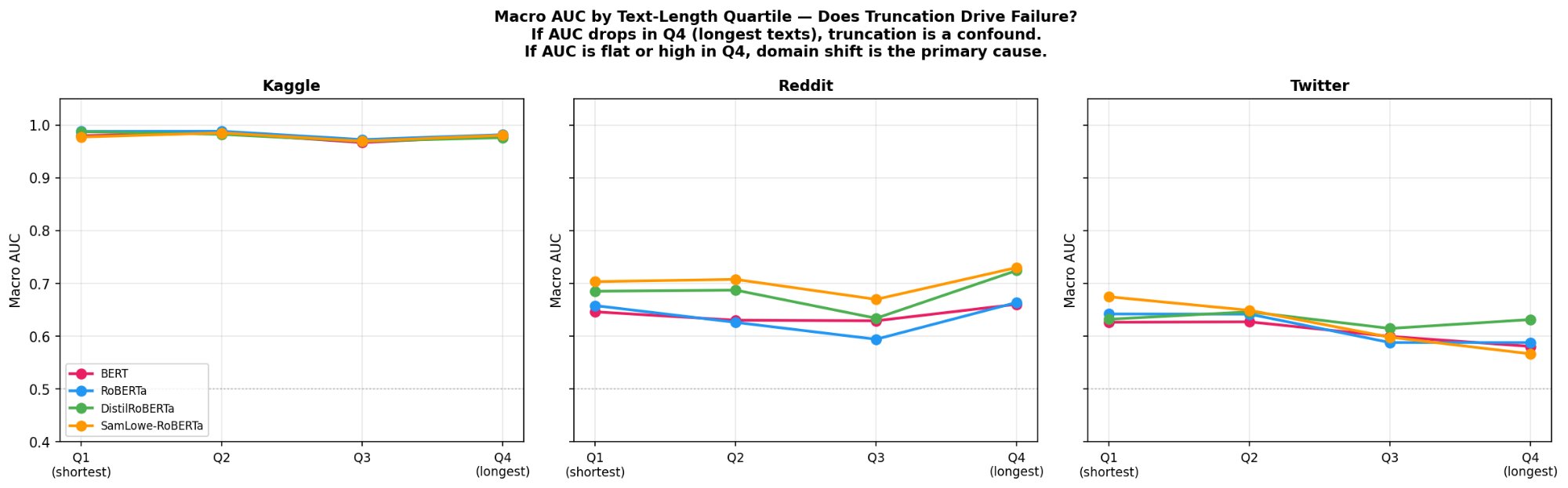}
  \caption{Supplementary Fig.\ S10: Macro \auc{} stratified by
    text-length quartile (Q1: shortest $25\%$; Q4: longest $25\%$).}
  \label{fig:s10}
\end{figure}

\begin{figure}[h!]
  \centering
  \includegraphics[width=\columnwidth]{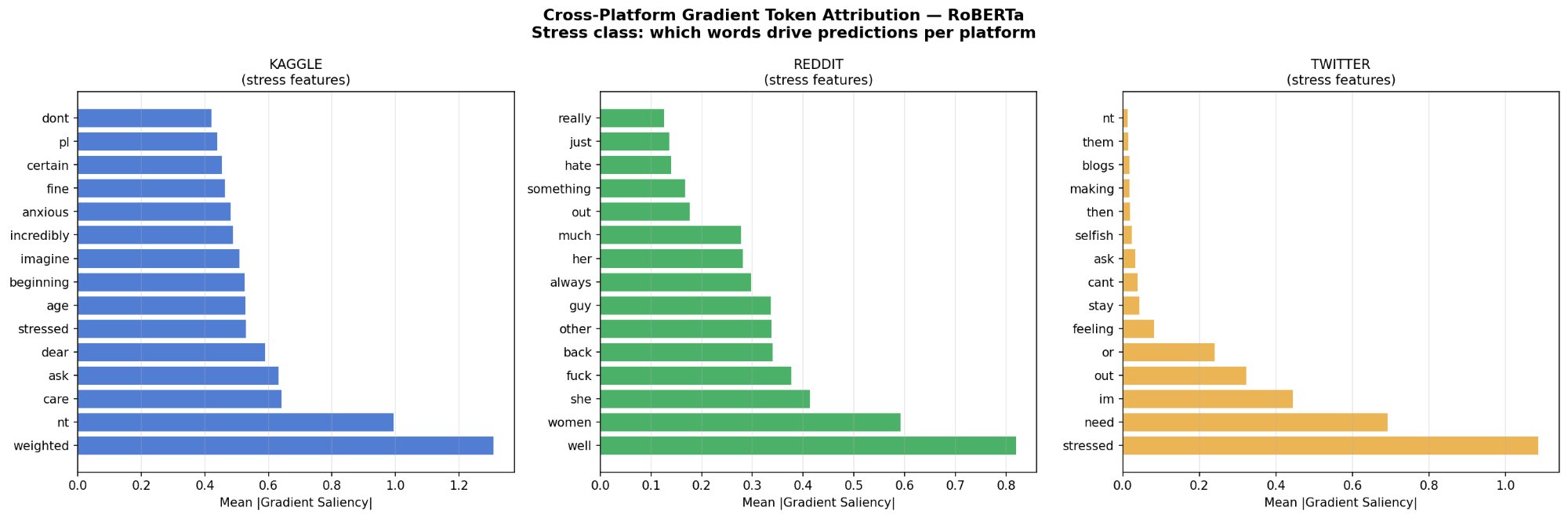}
  \caption{Supplementary Fig.\ S11: Cross-platform gradient token
    attribution for RoBERTa, Stress class.  Kaggle features include
    partially proxy-clinical tokens (\textit{weighted, stressed, care});
    Reddit and Twitter features shift to platform-specific conversational
    vocabulary.}
  \label{fig:s11}
\end{figure}

\begin{figure}[h!]
  \centering
  \includegraphics[width=\columnwidth]{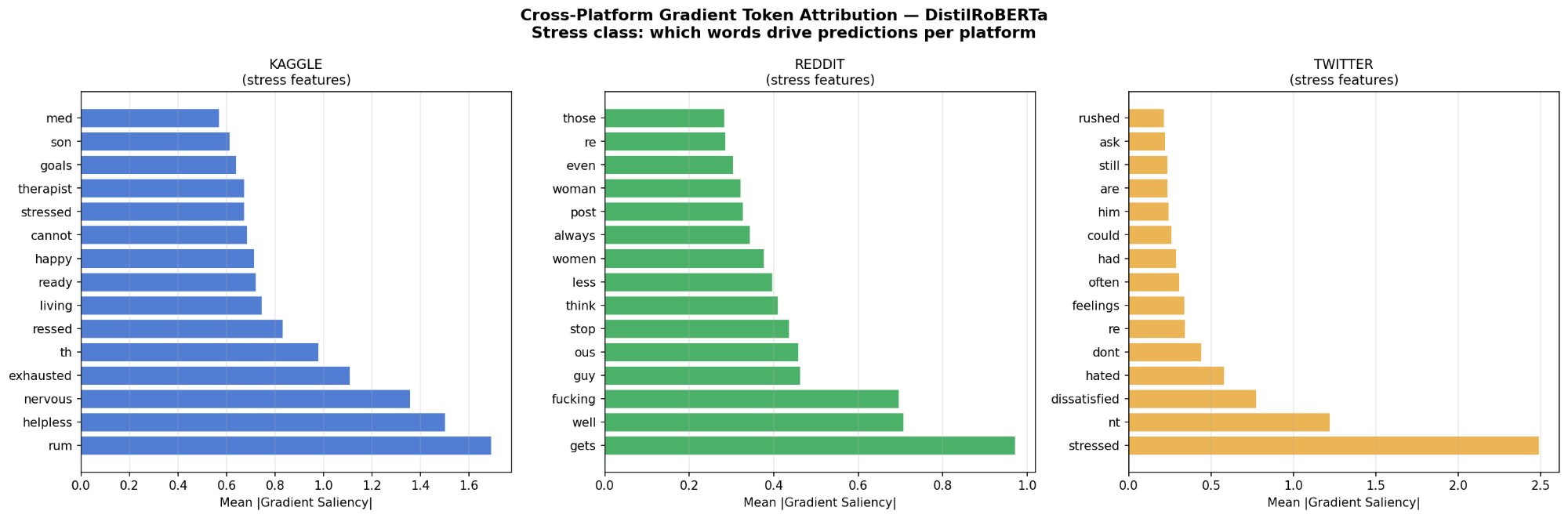}
  \caption{Supplementary Fig.\ S12: Cross-platform gradient token
    attribution for Emotion-DistilRoBERTa, Stress class.}
  \label{fig:s12}
\end{figure}

\begin{figure}[h!]
  \centering
  \includegraphics[width=\columnwidth]{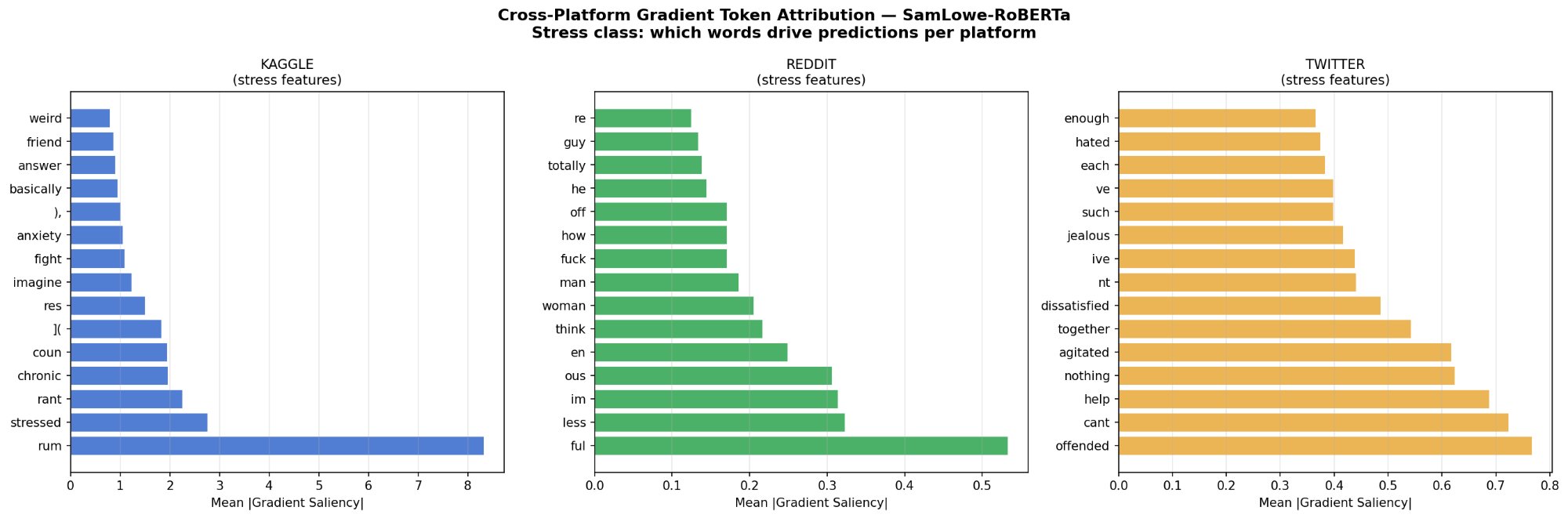}
  \caption{Supplementary Fig.\ S13: Cross-platform gradient token
    attribution for GoEmotions-RoBERTa, Stress class.  All three
    platform vocabularies are fully disjoint ($J{=}0.000$ for this
    model-class pair on Kaggle-to-Twitter comparison).}
  \label{fig:s13}
\end{figure}

\balance
\end{document}